\documentclass[11pt]{article}
\usepackage[preprint]{acl}

\usepackage{booktabs}
\usepackage{times}
\usepackage{amsmath}
\usepackage{hyperref}
\usepackage{cleveref}
\usepackage{latexsym}
\usepackage[T1]{fontenc}

\usepackage[utf8]{inputenc}

\usepackage{microtype}

\usepackage{inconsolata}

\usepackage{graphicx}

\usepackage{enumitem}

\usepackage{placeins}

\providecommand{\var}[1]{\textit{#1}}
\definecolor{promptboxborder}{rgb}{0.29,0.55,0.55}
\definecolor{promptboxheader}{rgb}{0.91,0.94,0.94}
\newsavebox{\pbbox}
\newenvironment{promptbox}[1][]{%
  \par\smallskip
  \begin{lrbox}{\pbbox}%
  \begin{minipage}{\dimexpr\columnwidth-2\fboxsep-2\fboxrule\relax}%
  \noindent\colorbox{promptboxheader}{%
    \makebox[\dimexpr\linewidth-2\fboxsep\relax][l]{\textbf{#1}}%
  }\par\smallskip\small\noindent\ignorespaces
}{%
  \end{minipage}%
  \end{lrbox}%
  \noindent\fcolorbox{promptboxborder}{white}{\usebox{\pbbox}}\par\smallskip
}

\title{Would you still call this Dax? \\ Novel Visual References in VLMs and Humans}

\author{
  \textbf{Ada Defne T\"ur\textsuperscript{$\heartsuit\clubsuit$}},
  \textbf{Gaurav Kamath\textsuperscript{$\heartsuit\clubsuit$}},
  \textbf{Joyce Chai\textsuperscript{$\spadesuit$}},
  \textbf{Siva Reddy\textsuperscript{$\heartsuit\clubsuit\diamondsuit$}},
  \textbf{Benno Krojer\textsuperscript{$\heartsuit\clubsuit$}}
\\
\\
  \textsuperscript{$\heartsuit$}McGill University,
  \textsuperscript{$\clubsuit$}Mila Quebec AI Institute,
\\
  \textsuperscript{$\spadesuit$}University of Michigan - Ann Arbor,
  \textsuperscript{$\diamondsuit$}Canada CIFAR AI Chair
\\
  \small{
    \textbf{Correspondence:} \href{mailto:ada.tur@mila.quebec}{ada.tur@mila.quebec}
  }
}

\begin{document}
\maketitle

\begin{abstract}
Vision-language models (VLMs), like human learners, are frequently exposed to new visual concepts, but how they map novel visual references to language after exposure remains largely underexplored, particularly when those references contradict prior knowledge from pre-training. To study this, we present the Novel Visual References Dataset (NVRD): 19,176 images spanning 90 visual concepts across different levels of visual novelty, each with up to 20 increasingly perturbed versions of the original object to probe generalization.
Unlike prior work on visual augmentations of familiar concepts, NVRD comprises entirely novel, open-ended stimuli constructed from scratch, mirroring how humans encounter genuinely new concepts. We evaluate 3 open- and 2 closed-source models alongside 2,400 human judgments for direct human–model comparison, and find that (i) models struggle to acquire novel concepts in-context when they contradict prior knowledge, and (ii) while models and humans show correlated sensitivity to visual perturbations, models significantly overgeneralize, extending learned labels to stimuli that humans reject. We contribute NVRD\footnote{\parbox[t]{\linewidth}{Code: \texttt{github.com/AdaDTur/nvrd}\\Data: \texttt{huggingface.co/datasets/adadtur/nvrd}}} as a corpus and benchmark for research on visual concept learning in both humans and machines. 
\end{abstract}

\section{Introduction}
\label{sec:intro}

As humans, we routinely encounter new objects and visual referents through our lifetimes---whether newly-invented technologies or culturally unfamiliar items (such as a \textit{paella} or a \textit{torii}).
We are also remarkable learners, and quickly adapt to such novel references with only a single or few instances of the referent; specifically, we apply certain biases acquired through previous knowledge to induce novel mappings between references and referents \citep{carey1978acquiring, markman1988children, merriman1989mutual}. For instance, humans, both children and adults, notably exhibit the shape bias: if an object's shape significantly changes, we are less likely to call it by the same name, compared to if only its color or texture changes \citep{landau1988importance,LANDAU1992807,landau1998object,samuelson2007dynamicnoun}.
Computational vision models, likewise, are regularly exposed to novel visual references at inference time, well after their initial training.
For instance, a user may present a model with an image of a newly-invented medical device, or a type of food that was not in its training data.
We ask: how do models categorize novel stimuli after exposure, and how well do they map them to their nonce names (\Cref{sec:multi-image}); do they generalize beyond perfectly identical instances of them (\Cref{sec:likert}); and how does this behavior compare with humans (\Cref{sec:human-study-main})?

\begin{figure*}[t]
    \centering
    \vspace{-6pt}
    \includegraphics[width=0.8\textwidth]{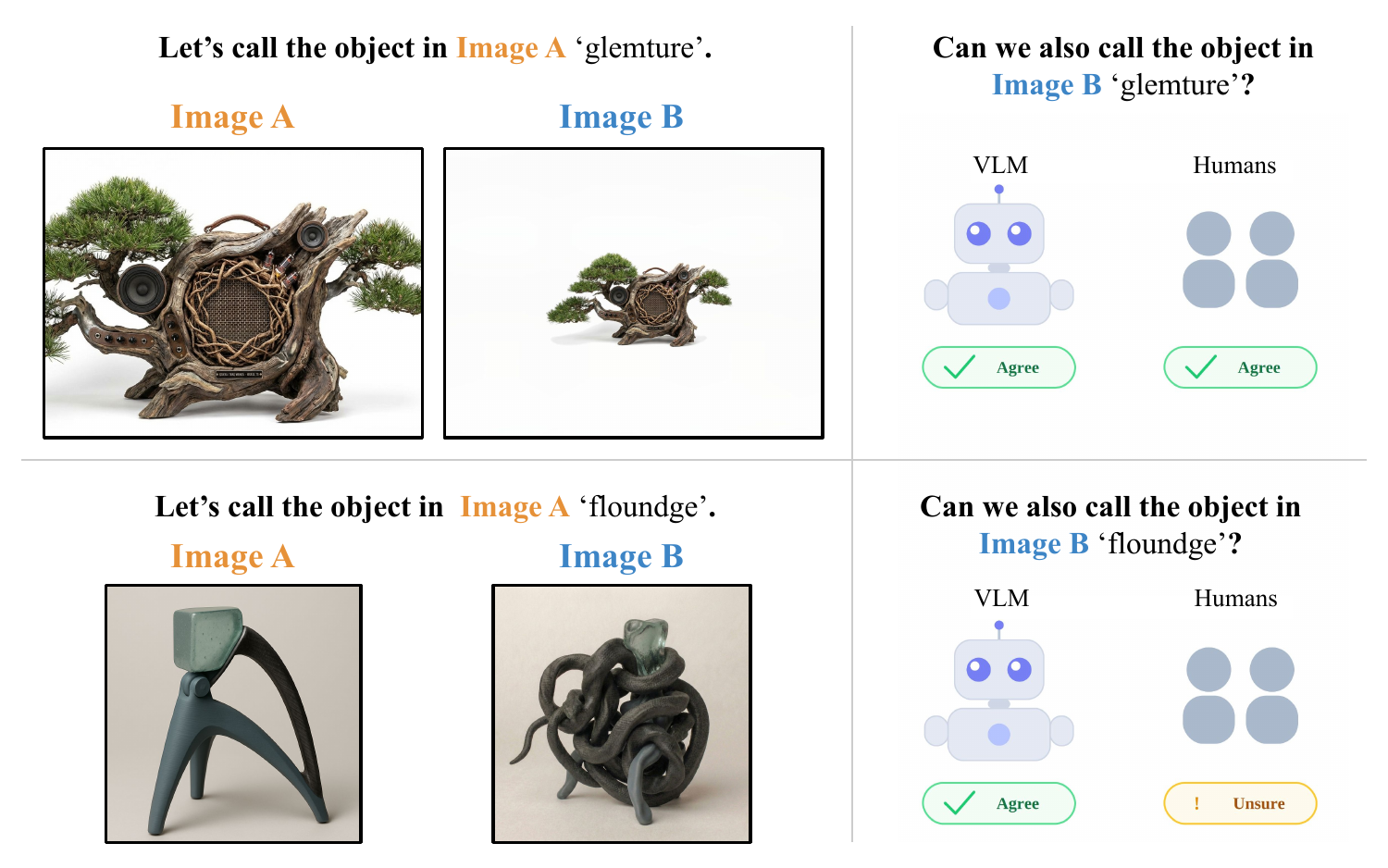}
    \vspace{-8pt}
    \caption{Task setup overview: On the left-hand side are examples of visual comparisons and nonce references evaluated; on the right-hand side, models and humans rate agreement with a novel label.}
    \label{fig:fig1}
    \vspace{-8pt}
\end{figure*}

To better understand how VLMs behave when exposed to novel visual concepts, we introduce the Novel Visual References Dataset (NVRD): 90 visual concepts spanning familiar objects (e.g. a lamp) to entirely new objects, each paired with a nonce word and up to 20 levels of controlled visual perturbations, totaling 19,176 images (see \Cref{fig:fig1}).
We evaluate three open-source VLMs\footnote{Our setup requires VLMs with multi-image capabilities.}: Qwen-2 VL 7B \citep{wang2024qwen2vl}, Idefics-3 8B \citep{laurencon2024buildingbetterunderstandingvisionlanguage}, and Molmo-2 8B \citep{deitke2024molmopixmoopenweights}, and two closed-source models: GPT-4o Mini \citep{openai2024gpt4ocard} and Gemini-2.5 Flash \citep{comanici2025gemini25}. Across three different prompting paradigms, we probe model behavior on NVRD under in-context learning (ICL) settings.
Doing so, we find that while models are capable of in-context acquisition of new visual concepts after exposure, this capability is reduced for stimuli that contradict prior conceptual knowledge.
To compare model behavior with humans, we then collect 2,400 human judgments on a subset of NVRD (see \Cref{fig:fig1} for an example trial).
While models tend to generalize novel learned concepts more widely than humans, both broadly agree on acceptability across perturbation types, scoring shape-based perturbations as less acceptable than texture or other low-level changes.
These findings connect to a broad literature on shape and texture biases in visual recognition \citep{geirhos2019texture, gavrikov2025talk}, while extending these questions to a novel-concept learning setting now made tractable by modern  image editing capabilities.

We release NVRD as a corpus with an interactive explorer for studying concept generalization in VLMs, and  hope it enables further cognitively-grounded evaluations of vision-language systems, as well as tools to probe how humans and agents communicate about a shared environment.

\section{Background}
\label{sec:background}

Human language acquisition has long been studied in linguistics, cognitive science, and philosophy.
\citet{quine1960word} proposed the inscrutability of reference: language learners theoretically face infinite potential mappings for any new word. Yet, children exhibit no such difficulty when acquiring language.

\citet{carey1978acquiring} and \citet{heibeck1987word} showed that children can form initial word-referent mappings after just one or two exposures---this is formally called fast-mapping---and \citet{SmithYu2008} and \citet{yu2007rapid} demonstrated that even infants use cross-situational statistics to spontaneously induce word-referent mappings in ambiguous contexts. Children further demonstrate \emph{learning biases} to constrain such potential mappings. For instance, the \textbf{shape bias}, children's tendency to rely on object shape rather than color, texture, or size to learn novel concepts, is among the earliest inductive biases exhibited for language acquisition \citep{landau1988importance, jones_smith_landau_1991, smith2002object, biederman_1987}. Children also demonstrate a \textbf{mutual exclusivity} bias, where they assume objects map to a single label, and thus assign novel words to unfamiliar objects \citep{markman1988children, markman1989categorization, merriman1989mutual}. These biases accelerate acquisition in language learners, particularly during early development, and carry on into adulthood \citep{LANDAU1992807}.

A related area of cognitive science research offers a theoretical perspective on human understanding of concept identity. Psychological essentialism is the hypothesis that humans represent categories as having hidden, causally deterministic identity that assigns membership independent of visual appearance, and is well-documented as a feature of conceptual structure in human perception \citep{medin1989conceptualstructure,gelman2003essentialism}. Empirical literature demonstrates that humans treat membership on an all-or-nothing basis, rather than graded and nuanced \citep{diesendruck1999domain}, use category membership over perceptual similarity for inductive inference from an early age \citep{gelman1986categoriesinduction}, and resist identity change under surface-level transformation in ways that artifacts do not \citep{keil1989concepts}. Under this view, visual similarity alone is insufficient to determine categorical boundaries; concepts are coherent to the extent that they are embedded in causal thoeries about the world \citep{Murphy1985TheRO}.

A central question in comparing human and machine vision is which visual features drive object recognition. \citet{geirhos2019texture} showed that ImageNet-trained CNNs are biased towards texture whereas humans favor shape; subsequent work argues this bias varies with architecture, training, and (for VLMs) language input \citep{gavrikov2025talk, hermann2020originsprevalencetexturebias}. These findings motivate our study of how visual biases manifest under novel concept learning.

A rapidly growing body of work investigates whether AI systems can acquire new concepts from limited exposure, mirroring fast mapping in humans \citep{carey1978acquiring, lake2015human, brown2020language}. Few-shot multimodal models such as Frozen \citep{tsimpoukelli2021frozen} and Flamingo \citep{alayrac2022flamingovisuallanguagemodel} learn new visual concepts in context; MEWL \citep{jiang2023mewl} reveals a large human--model gap under referential ambiguity, W2W \citep{ma-etal-2023-world} adds explicit grounding objectives for novel referents, and \citet{portelance2021shapebias} shows VLMs spontaneously adopt shape biases.
Novel word learning has also been studied in text-only settings via learning procedures \citep{hewitt2025neologism, wang2025rapid} and inference-based probing \citep{brubaker-etal-2026-wugnectives}.
These contributions, however, mostly evaluate simplified or known object classes (e.g., MEWL combines known shapes/colors; Flamingo uses standard benchmarks); we extend evaluation to truly novel visual referents.

\section{The Novel Visual References Dataset}
\label{sec:nvrd}

\subsection{Dataset Overview}
\label{sec:dataset-overview}
\begin{figure*}[t]
    \centering
    \vspace{-6pt}
    \includegraphics[width=\textwidth]{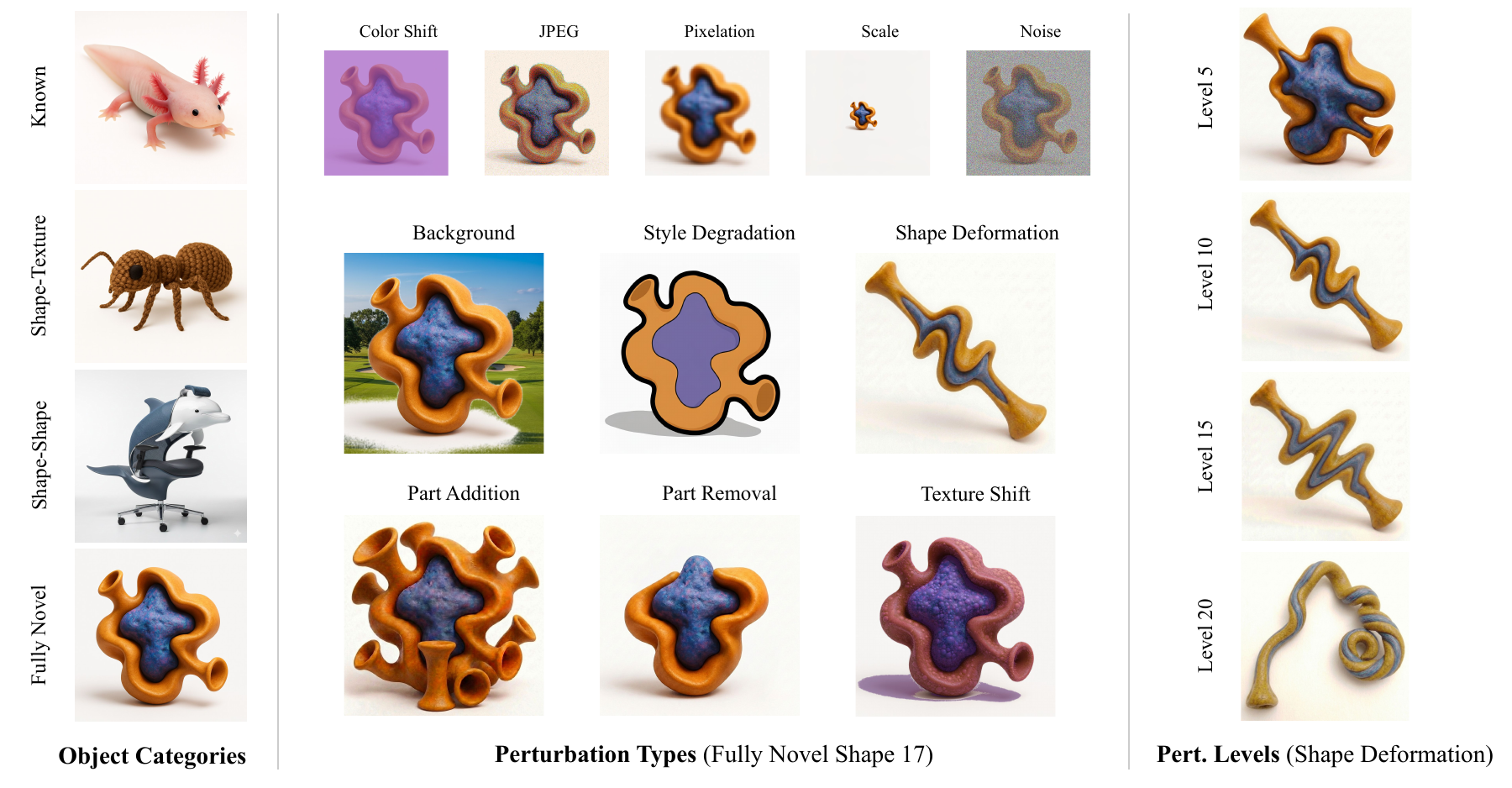}
    \vspace{-8pt}
    \caption{Overview of the creation pipeline for NVRD. On the left-most column, we present examples of the four object categories; in the middle column, we show all 11 perturbation axes on one example fully novel stimulus, enlarging the high-level edits; in the right-most column, we present how the shape deformation perturbation modifies the novel object shape monotonically over 20 levels.}
    \label{fig:fig2}
    \vspace{-8pt}
\end{figure*}

We introduce the Novel Visual References Dataset, or NVRD, a corpus of 90 images of unique objects and entities ranging across a spectrum of novelty and different types of compositions, plus an additional set of perturbations for each image across 11 augmentation axes, totaling 19,176 unique images. Base images are generated using Gemini-3 Pro Image, and perturbations are produced using both Gemini-3 Pro Image and Gemini-2.5 Flash Image; we present a summary of the dataset curation \Cref{fig:fig2}. All generated images undergo a two-stage automated quality control pipeline followed by manual validation, with full generation prompts and procedures described in App.~\ref{sec:generation}.

We begin by organizing visual concepts into three categories: \textit{known}, \textit{composed} and \textit{fully novel} entities.
Whereas fully novel entities aim to represent objects that do not exist or resemble anything that exists in the real world, composed entities combine specific attributes and components of existing objects; these are further distinguished between shape-shape compositions (e.g. a boar-toaster hybrid), and shape-texture compositions (e.g. a lion with bird feathers). Known entities depict objects which commonly occur in the model's training (e.g. a chair); however, these too are given nonce names (e.g. the chair is given the name \emph{"blomwich"}), allowing us to study cases that contradict prior conceptual reference knowledge.
We generate 30 known entities, 30 composed entities, and 30 fully novel entities, totaling 90 base objects (see App.~\ref{sec:image-gen-settings} for the full list of concepts and generation prompts).

\subsection{Image Perturbations}
\label{sec:perturbations}
To study how visual augmentations affect model concept judgments, we construct controlled perturbation sequences for each object in our dataset. A major consideration is that not all visual modifications are equal; adding noise to an image is categorically different from deforming an object's shape. We therefore select a range of visual perturbations motivated by distinct findings from model robustness benchmarks, cognitive science, and representation learning, which, together, span multiple dimensions along which stimuli can visually vary \citep{hendrycks2019robustness,hendrycks2021facesrobustnesscriticalanalysis}.

Let $x_0$ denote an image from our dataset, and let $\mathcal{P} = \{p_1, \dots, p_{11}\}$ be the set of perturbations. For each perturbation $p \in \mathcal{P}$ we construct a \emph{compounding sequence} of $L$ levels: 

\begin{equation}
 x_0 \;\xrightarrow{p}\; x_1 \;\xrightarrow{p}\; x_2 \;\xrightarrow{p}\; \cdots \;\xrightarrow{p}\; x_L
\end{equation}

where each $x_\ell$ is generated by applying $p$ to $x_{\ell-1}$. Because the same perturbation is re-applied to its own output, the visual distance from $x_0$ ideally increases monotonically with $\ell$, giving a continuous axis of perturbation intensity along which we can study patterns of model judgments on novel concepts; we verify this compounding using a set of quality control procedures, described in App.~\ref{sec:generation}, in addition to a manual author validation (see App.~\ref{sec:app-author-val}) on 100 random datapoints with high-level perturbations, of which 18\% were noise/undesirable. We determine $L$ based on each image independently, as certain perturbations tend to \textit{saturate}\footnote{Saturation is determined by the VLM judge, detailed in App.~\ref{sec:generation}}, which we note for relevant perturbations; we select 11 perturbation axes, which we describe and motivate as follows.

\subsubsection{Low-Level Edits.}
We first use five standard image corruptions that alter low-level visual properties without modifying the object's shape or structure: \textbf{Gaussian Noise} ($\epsilon \sim \mathcal{N}(0, \sigma^2 \mathbf{I})$), \textbf{Scale}, which we produce with simple zooming augmentations, and \textbf{Pixelation} (nearest-neighbor down-sampling) allow us to measure the spatial granularities and noise ratios at which models still extend mappings, drawing on the finding that humans can recognize objects from highly reduced patches \citep{ullman2016atoms}, whereas models often rely on local texture \citep{geirhos2020generalisationhumansdeepneural}. \textbf{JPEG Compression} introduces color banding and artifacts, which allows us to probe the effect of high-frequency information loss \citep{dodge2016understandingimagequalityaffects}. Finally, we apply \textbf{Color Shift}, which we conduct by applying an arbitrary hue filter of increasing intensity, allows us to probe color bias.

\subsubsection{Higher-level Edits}
\label{sec:highlevel}
We produce higher-level edits using Gemini-3 Pro Image and Gemini-2.5 Flash to apply perturbations to a source image, detailed further in App.~\ref{sec:image-gen-settings}.

\textbf{Texture Shift:} We transfer a target texture (e.g., a slime-like surface) onto an object while preserving its shape, then linearly interpolate between the original and textured images across 20 levels. Texture plays a central, though contested, role in object recognition \citep{geirhos2019texture, hermann2020originsprevalencetexturebias}; we examine its effect alongside shape and color perturbations.

\textbf{Background:} We generate a contextually inappropriate background (e.g. a pub scene behind a humpback whale) and increase its opacity across 20 levels, probing whether models exploit context as a classification shortcut \citep{beery2018terra, xiao2021backgrounds}.

\textbf{Artistic Style:} We generatively apply a progressively rougher rendering style (saturating after $\sim$16.3 levels on average), probing whether models rely on fine-grained rendering cues rather than structural form.

\textbf{Shape Deformation:} We generatively deform the object's silhouette and geometry across 20 levels---the perturbation expected to most directly erode concept identity given the centrality of shape in human and machine learning (\Cref{sec:background}).

\textbf{Part Addition:} We generatively append an extra part or limb across 20 levels; the compounding sequence ensures level $L$ has at least $L$ extraneous parts, probing compositional understanding \citep{ma2023crepe}.

\textbf{Part Removal:} Each level removes a part, limb, or appendage (saturating after $\sim$16 levels on average), probing how learning degrades when parts---shown to serve as recognition cues in humans \citep{biederman_cooper_1991}---are removed.

\section{Experiments}
\label{sec:experiments}

Given that VLMs can learn new concepts through a variety of settings (in-context learning, pre-training, post-training, etc), we consider which approach is most faithful to our goal of understanding how VLMs handle and acquire new visual concepts "in the wild"; in-context learning aligns more closely with how adults instinctively acquire novel vision-language mappings without repeated exposure and instruction.
Since purely prompting-based approaches raise questions of linguistic faithfulness \citep{hu2023promptingsubstituteprobabilitymeasurements}, we use three separate behavioral probing methods on the five models listed in \Cref{sec:intro}, which we describe and motivate below.
Throughout all experiments, models are exposed to a base image $x_0$ of a novel object paired with a nonce word $r$, and a perturbed variant $x_\ell$ at perturbation level $\ell$. The specific inputs and elicitation method vary across the three setups described below.

\subsection{Name Generation from Multi-Image In-Context Learning}
\label{sec:multi-image}
In our first probing set-up, each visual stimulus $x_0$ is paired with a nonce word $r$, constructed by prompting GPT-4o to generate candidates and filtering for nonce words with exactly three tokens in length.
We build an in-context pool $\mathcal{C}$ consisting of $x_0$ captioned with $r$, four distractor image-caption pairs (the most visually similar images to $x_0$ from a pool of 20,000 PixMoCap images \citep{deitke2024molmopixmoopenweights} using CLIP ViT-B/32 \citep{radford2021learning}, each with a single-word caption), and $x_\ell$ with a fill-in-the-blank caption: ``This image is best described by the reference: \_\_\_\_.''
We shuffle the full image pool and always show $x_\ell$ last. All models use greedy decoding and generate responses to fill the blank, re-generating up to three times if the response is shorter than 2 characters; prompting details are provided in App.~\ref{sec:experimental-details}.

\subsection{Token Probabilities Given Multi-Image In-Context Learning}
\label{sec:tokprob}
As previously mentioned, prompting may not always faithfully represent model preferences.
We therefore additionally probe the three open-source models for the log probabilities they assign to our nonce words for each image. Using an identical setup to the multi-image generation, we present the shuffled image pools to the models, but, rather than a fill-in-the-blank task, we provide the final image caption with the target nonce reference and we compute the reference probability using the following formulation:

\vspace{-10pt}
\begin{equation}
    \frac{1}{N}(log P(\textcolor{blue}{r} \mid \mathcal{C})) = \frac{1}{N}(\sum_{i=1}^{N} \log P(t_i \mid t_{<i}, \mathcal{C}))
\end{equation}
\vspace{-10pt}

where $\textcolor{blue}{r}$ is our target nonce reference, $N$ is the token-length of the reference, $t_1, t_2, \ldots, t_N$ are its constituent tokens, and $\mathcal{C}$ is the full in-context image pool with captions, the instruction, and the final prompt containing the target caption. We compute $\log P(t_i \mid t_{<i}, \mathcal{C})$ by applying log-softmax over the model's output logits and selecting the entry for $t_i$. We also compute the probability of "vanilla" references (e.g. "tree frog" instead of the assigned nonce label for an image of a tree frog), to compare whether our models are genuinely acquiring the novel mappings, or defaulting to labeling using familiar concepts. This nonce--vanilla contrast controls for pure recency or continuation effects from the in-context label: such effects would assign comparable probabilities to both.

\subsection{Dual-Image Likert-Scale Rating}
\label{sec:likert}
Finally, as our third experimental setting, we use a dual-image Likert-scale rating setup, where we only present the models with $x_0$ and $x_\ell$, excluding any other distractors. The model is first shown $x_0$ captioned ``Let's call the object in this image `[nonce word]'.'' Then, it is shown $x_\ell$ and asked to rate agreement with the statement ``Could both of these images be called `[nonce word]'?'' on a scale from 1 to 7, where 1 = Strongly Disagree and 7 = Strongly Agree. The model responds with a single integer which we parse and collect.
This set-up aligns closely with the experimental set-up we use for our human study (see below); we use model results from this setting for fair model-human comparisons.

\subsection{Human Study}
\label{sec:human-study-main}
Finally, we compare human and model judgments around our visual stimuli and novel references under the same Likert-Scale-based experimental setup.
We conduct a crowd-sourced study through Prolific, collecting judgments from 30 anonymous native English speakers on 800 unique image pair trials. We describe participant details, compensation, and privacy in App.~\ref{sec:human-study}.
While this represents a subset of NVRD, we focus on the higher-level perturbation types discussed in \Cref{sec:perturbations} which are most motivated by cognitive findings, and our sampling design ensures each image pair receives multiple independent judgments to support reliable mean rating estimates.
\Cref{fig:study_ex} shows the set-up of our human study. Each participant sees 80 image pairs and each image pair receives three separate human judgments, totaling 2400 human judgments. On each trial, participants see the original image and a perturbed version side-by-side, along with one of our nonce words; they are asked the same prompt as models, and to respond on a 7-point scale from "Strongly Disagree" to "Strongly Agree."

\section{Results \& Discussion}
\label{sec:results}
We present our primary results and discussion in the following sections, with more results in App.~\ref{sec:additional-results}.
\paragraph{Models acquire novel references in-context, but struggle when they conflict with prior knowledge.}

\begin{figure*}[h]
    \centering
    \includegraphics[width=\textwidth]{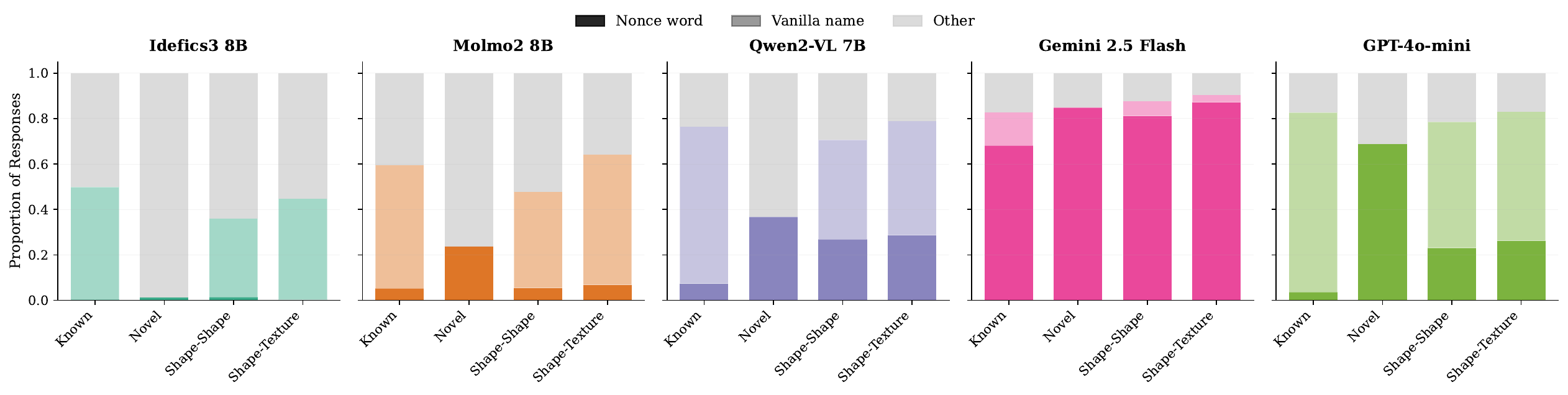}
    \caption{Nonce vs. vanilla label responses across models and object categories. We find: Models adopt a nonce words most easily for novel or partially novel objects.}
    \label{fig:nonce-vanilla-obj-cat}
\end{figure*}

\begin{figure*}[h]
    \centering
    \includegraphics[width=\textwidth]{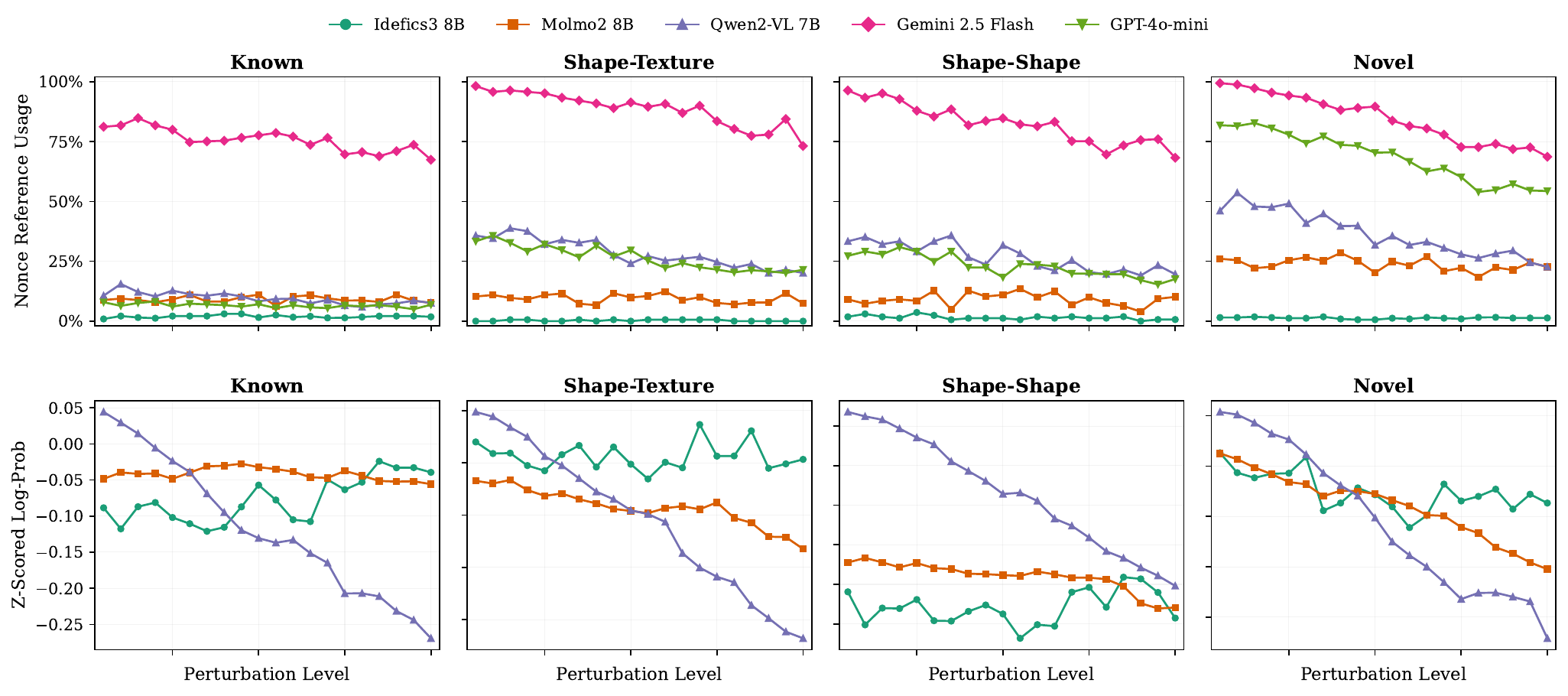}
    \caption{Model results on both the multi-image name generation and log probability settings across object categories. Log probabilities are $z$-scored to make them comparable across models.}
    \label{fig:ref-gen-prob-obj-cat}
\end{figure*}

When examining nonce usage across object categories in the name generation setup (\Cref{sec:multi-image}), we observe a noticeable effect of object novelty (\textit{known}, \textit{composed}, \textit{novel}), which we present in \Cref{fig:nonce-vanilla-obj-cat}.
Known entities consistently receive the lowest nonce reference usage across all models, with GPT-4o Mini displaying a particularly strong preference towards existing ``vanilla labels''\footnote{Known objects have one vanilla label (e.g. ``dog'' for dog), composed objects have \textit{two} such labels (e.g. ``dog'' and ``cat'' for a hybrid object of the two), novel objects have none.} instead of novel learned mappings. This echoes \emph{mutual exclusivity} effects \citep{merriman1989mutual,markman1988children}, where learners are less willing to apply a new label to an object that already has an existing mapping. Novel objects, on the other hand, see the highest nonce reference usage, suggesting that the absence of competing known labels lowers the threshold for models to commit to using novel labels, while composed entities fall in between the two, with shape-texture compositions seeing slightly more nonce usage than shape-shape compositions.
Closed-source models respond with slightly less hesitation overall, with Gemini-2.5 Flash exhibiting the highest nonce reference usage across all models tested.
Across perturbation levels, models which do adopt the target nonce reference tend to continue doing so as visual distance from $x_0$ increases, though with declining rates (\Cref{fig:ref-gen-prob-obj-cat}). This suggests that a single in-context exposure is sufficient to sustain some level of generalization, even as visual distance from the original increases. However, for some models (e.g. Idefics3 or Molmo2) generation curves are mostly flat, making it difficult to assess how confidence in novel references changes with perturbation. We therefore additionally examine log probabilities.

\paragraph{Log probabilities reveal more nuanced trends}
\label{sec:5.2}


In the more nuanced log probability setup (\Cref{sec:tokprob}), we find that across all three open-source models, log-probability declines with perturbation level---steepest for Qwen-2 on part removal ($z$-scored drop of 0.8). Across object categories, log probabilities largely mirror the generation results but sometimes show more gradual trends. Known entities show flat, low nonce log-probabilities. Shape-shape compositions show a clearer decline than shape-texture, suggesting structurally hybrid objects are more sensitive to further structural perturbation.

\begin{figure*}[h]
    \centering
    \includegraphics[width=\textwidth]{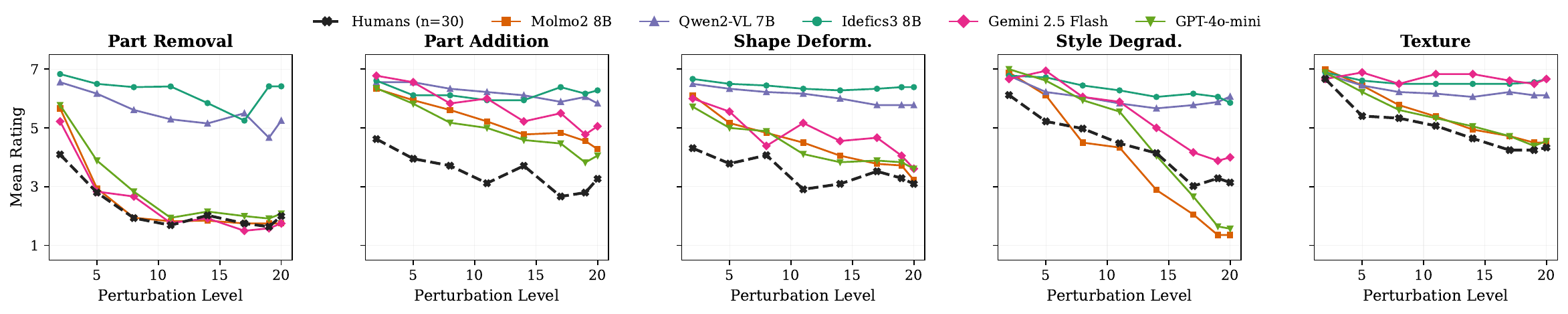}
    \caption{Human and model ratings on the subset of perturbation types that show a clear degradation at strong perturbation levels (i.e. high-level edits from \Cref{sec:highlevel}, except background)}
    \label{fig:human-model-pert-type-subset}
\end{figure*}

\paragraph{When generalizing novel concepts, models are most sensitive to structural perturbations (e.g. shape).}
In the dual-image Likert-scale setup (\Cref{sec:likert}), we focus on perturbation types (shape, texture, etc.) rather than object categories. This setup is particularly suited for this analysis for two reasons: it enables a fair comparison with human judgments (\Cref{sec:human-study-main}), while sidestepping the issue that some models struggle to reliably adopt novel references in the previous setting, making Likert ratings a more direct signal of concept identity judgments. In \Cref{fig:human-model-pert-type-subset}, VLMs rate whether they fully agree (7) or disagree (1) that a perturbed object should be assigned the same nonce word as the original one:
Here, even at perturbation level 1, model judgment starts slightly at lower ratings of 6 (\textit{Somewhat Agree}/\textit{Agree}) for shape-related perturbations (part removal, part addition, shape deformation).
At stronger perturbation levels, particularly part removal drops very low to ratings between 1 and 2.
Interestingly, some models (GPT-4o-mini, Molmo2) are also sensitive to texture to a lesser extent.
Other less semantic perturbations (\textit{low-level edits}) such as resizing the object or color shift have almost no effect with model judgements remaining close to 7 (\textit{strongly agree}); the only exception is Molmo2-7B that assign lower ratings below 5 for color perturbations (detailed breakdown in \Cref{app:g3}).
Overall, we can confirm existing findings \citep{gavrikov2025talk} that modern VLMs are more shape-biased, now in a more realistic scenario with novel objects and perturbations, and also find some influence of texture.

\paragraph{Models and humans strongly correlate on novel reference generalization across perturbation types, but models over-generalize.}
\label{sec:5.3}

A central question motivating our study is not only whether VLMs can acquire novel visual references, but also how their generalization patterns compare with those of human learners. To conduct this comparison, we analyze VLM Likert ratings against judgments from 30 human participants using the same dual-image task, finding both meaningful agreements and divergences.

We first find that humans and models are broadly aligned in their sensitivity to different perturbation types, as shown in \Cref{fig:human-model-pert-type-subset}. Human participants show the steepest declines in ratings for part addition and removal, followed by shape deformation and style degradation---an ordering identical to that observed across models. This alignment suggests that both humans and VLMs have internalized something of a shape bias in the context of novel concept generalization, treating modifications to an object's structural configuration as more identity-threatening than surface-level changes to color, texture, or background. Spearman correlations by leave-on-out between human and model ratings across perturbation types further confirm this agreement, where correlations are weakest for Idefics-3 8B, ranging from $\rho$ = 0.688 for Background and $\rho$ = 0.770 for Part Removal, and strongest for GPT-4o Mini, ranging from $\rho$ = 0.899 and $\rho$ = 0.925.

Despite this broad agreement, a clear difference emerges when we examine object categories. Human participants are substantially more influenced by object novelty, rating novel entities on average two full points lower than known entities by the final perturbation level; model ratings, by contrast, are relatively flat across all object categories (detailed breakdown in \Cref{fig:human-model-obj-cat}).

This is most noticeable when comparing raw rating distributions, where humans produce the lowest mean ratings across all conditions and exhibit significantly more variability, while models cluster near the upper end of the scale with compressed variance, particularly the open-source models. The gap is most severe for shape-based perturbations, where human ratings fall below 3 (\textit{Somewhat disagree}) by perturbation level 10 on average for part removal and shape deformation, while most models remain between 4 (\textit{Neutral}) and 6 (\textit{Agree}) at the same level; we ablate on this over-generalizing behavior in \Cref{sec:syco-ablation}. For texture and color shift, by contrast, humans and models are closely aligned, suggesting genuine agreement on which perturbation types are most semantically significant. Thus, the divergence is more so that models recognize a similar hierarchy of perturbation severity that humans do, but apply it with far less discrimination, extending novel labels to stimuli that we would reject.

The category-sensitivity gap we observe, that humans rate novel entities considerably lower than known ones as the perturbation level increases, while models remain relative flat across categories, is consistent with the essentialist account of human categorization \citep{gelman2003essentialism,keil1989concepts}, where humans treat established categories as supported by properties that surface-level changes cannot alter, while models appear to predominantly track visual similarity.

\section{Ablations}
\label{sec:ablations}

We ablate the role of CLIP visual similarity on model ratings and the metric used to compose the in-context image pool; both confirm that our visual-similarity setup is the least trivial task (App.~\ref{app:vis-similarity-ablation}, App.~\ref{app:pool-composition}). We focus the main-body discussion on a behavioral ablation that probes the over-generalization we observe.

\subsection{Prompt Agreement Ablation}
\label{sec:syco-ablation}
Since we notice that many of the models appear to over-generalize learned concepts, we also probe whether this over-generalization is driven by a broader pattern of ``agreeing with everything'' shown to the model.
On Qwen2-VL, we re-run a subset of 1000 randomly sampled image pairs from NVRD under the Likert-scale rating experimental setup, but make sure that the image pairs in the prompt are entirely different objects from one another.
For example, even a fully novel object could now be paired with a known object and the model is asked whether they would assign the same reference.
Any human would assign 1 or 2 (\textit{disagree}) to such examples.
However model now most often assign 3 (\textit{somewhat disagree}), and in 30\% of the cases even assign 6 (\textit{agree}).
We further find that most of these assigned 6 judgments occur when the second images comes from the \textit{fully novel} category.
We can conclude that while models and humans have high agreement in our main experiments, they diverge in this more adversarial setup.
Our observation aligns with the performance observed in the multi-image in-context learning settings, where Qwen-2 utilizes the target nonce reference much more when they are mapped to novel entities.
Details in \Cref{sec:additional-ablations}.

\newpage

\section{Conclusion}
\label{sec:conclusion}
In this work, we investigated how VLMs learn and generalize novel visual concepts, and how their behavior compares with human judgments. Evaluating five VLMs across three prompting paradigms on our Novel Visual References Dataset (NVRD), we find three key results. First, while models are capable of acquiring novel visual references in context, this capability is undermined when new stimuli conflict with prior conceptual knowledge, echoing mutual exclusivity effects well-documented in human word learning. Second, models proved most sensitive to shape-based perturbations like part addition and removal, whereas surface-level changes like color shift and background had almost no effect, suggesting models track something of a shape bias, even in purely novel referential contexts. Third, while models and humans strongly correlate in their overall sensitivity to perturbation type, models over-generalize and extend novel labels to heavily perturbed stimuli that human participants would reject.

This misalignment has direct implications for reliable human-agent interaction, where a human and agent must jointly establish and maintain shared referents for objects in a common and continuously evolving environment. Models that show a strong asymmetry between accepting and producing novel labels will struggle to act as consistent and capable communicative partners in such settings. We contribute NVRD as an open-source corpus and evaluation tool to encourage further research into such cognitive gaps, and towards the development of vision-language systems that can learn, generalize, and communicate about novel visual concepts with the sensitivity and grounding we observe in human learners.

\newpage
\section*{Limitations}
\label{sec:limitations}
NVRD is constructed using state-of-the-art generative image models (Gemini-3 Pro Image and Gemini-2.5 Flash Image), which themselves embed biases over object structure, materials, and visual style. Our manual author validation on 100 high-level perturbation pairs found 18\% to be noisy or undesirable, and several perturbation types saturate before reaching their nominal 20 levels. Although our two-stage VLM judge and post-hoc cleaning mitigate these effects, residual generator artifacts are likely present, particularly for compositional and fully novel categories where the generator has weaker priors.


\section*{Ethical Considerations}
\label{sec:ethics}
Our human study was conducted via Prolific with 30 anonymous adult participants, all native English speakers residing in the US, Canada, UK, or Ireland. Participants were paid an average of \pounds17.79/hour, were informed of how their responses would be used and of their rights regarding submitted data, and could withdraw at any time. No personally identifying information was collected, and we report only aggregate statistics. Full details are provided in Appendix~\ref{sec:human-study}.

\section*{Acknowledgments}
We would like to thank Eva Portelance and Jeonghwan Kim for the interesting and helpful conversations and feedback during our research. This work was made possible by funding from the IVADO R3 NLP Régroupement.

\newpage

\bibliography{custom}
\newpage
\appendix
\section*{Appendix: Table of Contents}
\label{sec:appendix-toc}

\begin{itemize}[leftmargin=1.5em, itemsep=2pt]
    \item[\textbf{A}] \nameref{sec:dataset-examples} \dotfill \pageref{sec:dataset-examples}
    \item[\textbf{B}] \nameref{sec:image-gen-settings} \dotfill \pageref{sec:image-gen-settings}
    \item[\textbf{C}] \nameref{sec:gen-prompts} \dotfill \pageref{sec:gen-prompts}
    \item[\textbf{D}] \nameref{sec:generation} \dotfill \pageref{sec:generation}
    \item[\textbf{E}] \nameref{sec:experimental-details} \dotfill \pageref{sec:experimental-details}
    \item[\textbf{F}] \nameref{sec:human-study} \dotfill \pageref{sec:human-study}
    \item[\textbf{G}] \nameref{sec:human-model-correlations} \dotfill \pageref{sec:human-model-correlations}
    \item[\textbf{H}] \nameref{sec:additional-ablations} \dotfill \pageref{sec:additional-ablations}
\end{itemize}

\newpage
\section{Dataset Examples}
\label{sec:dataset-examples}

\begin{figure*}[!t]
    \centering
    \includegraphics[width=0.7\textwidth]{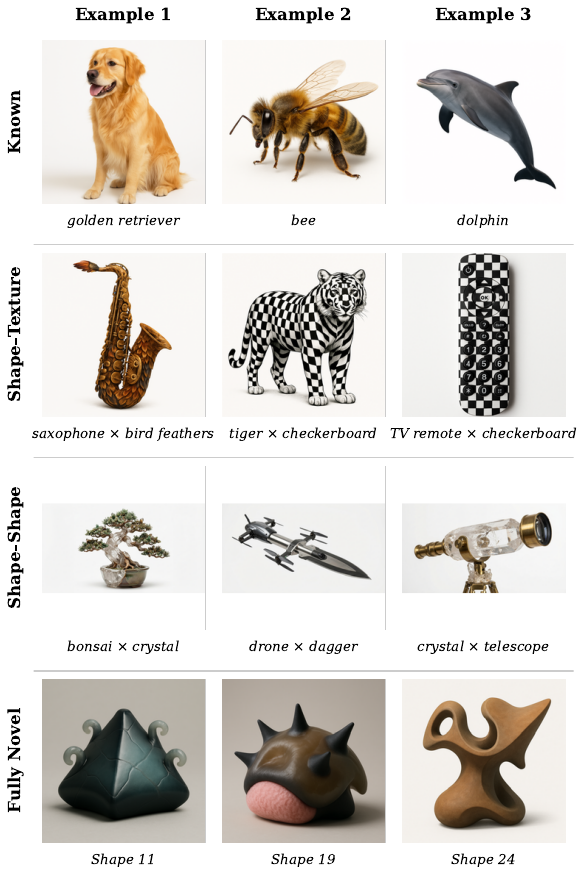}
    \caption{Example base images from each of the four entity categories in NVRD. \textbf{Known} entities are familiar objects likely present in pre-training data. \textbf{Shape--Texture} compositions combine a known object's shape with a different texture. \textbf{Shape--Shape} compositions merge two known objects into a single cohesive entity. \textbf{Fully Novel} entities are designed from scratch and do not correspond to any real-world object.}
    \label{fig:base-examples}
\end{figure*}

\begin{figure*}[!t]
    \centering
    \includegraphics[width=\textwidth]{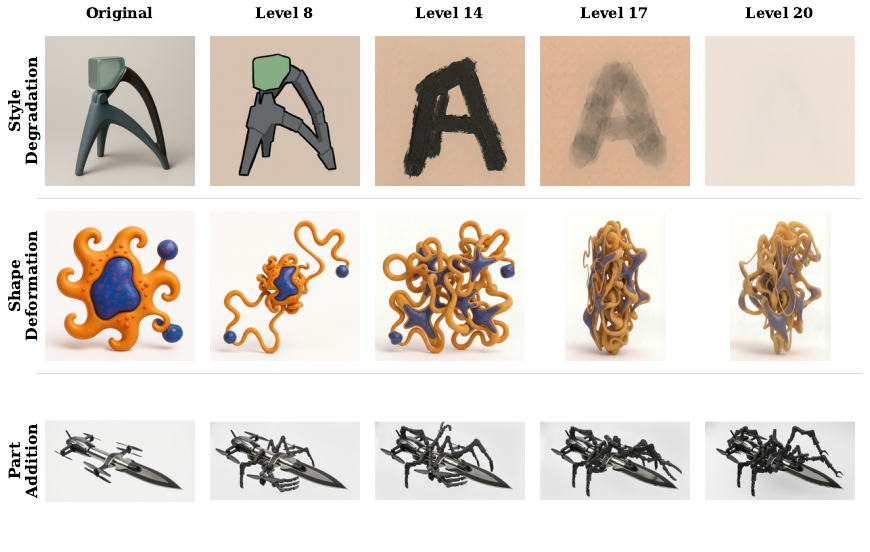}
    \caption{Example perturbation sequences from NVRD. Each row shows an original base image and four increasingly perturbed variants along a single perturbation axis. \textbf{Top:} Style degradation applied to a fully novel entity, progressively reducing artistic fidelity. \textbf{Middle:} Shape deformation applied to a fully novel entity, warping and distorting the object's silhouette. \textbf{Bottom:} Part addition applied to a shape--shape composition (drone $\times$ dagger), progressively appending new structural elements.}
    \label{fig:pert-examples}
\end{figure*}

This section provides visual examples from NVRD to illustrate the range of entity categories and perturbation types in the dataset. \Cref{fig:base-examples} shows example base images spanning all four entity categories, and \Cref{fig:pert-examples} shows three representative perturbation sequences applied to base images from the dataset.

\newpage
\section{Image Generation Prompts and Settings}
\label{sec:image-gen-settings}

All base images are generated using Gemini-3 Pro Image \citep{comanici2025gemini25}. We observed that the creation of novel entities, as well as their perturbations, was disproportionately more trivial for generative models compared to all other entities, whereas known entities were the most difficult. We hypothesize this is due to the model's motivation to preserve the original semantic composition of an object it is already familiar with from its training data, thus it struggles to envision unique perturbations to familiar concepts. Below we list the object pools and generation prompts for each of the four entity categories. Visual examples of base images and perturbation sequences are provided in App.~\ref{sec:dataset-examples}, and a full overview of the human study sample is shown in \Cref{fig:study-sample-grid}.

\subsection{Known Entities}
\label{app:known-entities}

\begin{promptbox}[Known Characteristics]
\small
\textbf{Objects:} ant, axolotl, backpack, bald eagle, bat, bear, bee, black widow, bow tie, camel, chair, chimpanzee, clock, coffee table, corkscrew, crocodile, dolphin, fedora, fire alarm, golden retriever, horse, humpback whale, ladle, leopard shark, moose, rabbit, siamese cat, tree frog, turtle, wallet
\end{promptbox}

\begin{promptbox}[Generation Prompt]
\small
\texttt{"Generate an image of a(n) \var{obj}. Make the background white and do not add anything else. Use a realistic art-style."}
\end{promptbox}

\subsection{Shape-Texture Compositions}
\label{app:shape-texture}

\begin{promptbox}[Shape-Texture Characteristics]
\small
\textbf{Shapes:} ant, headphones, horse, ladybug, lion, saxophone, bee, camel, moose, panda, tiger, TV, broom, coyote \\

\textbf{Textures:} bird feathers, checkerboard, crochet
\end{promptbox}

\begin{promptbox}[Generation Prompt]
\small
\texttt{"Generate an image of a realistic <object>, but with the following <texture> texture: <image of texture>. Make the background white and do not add anything else to the image."}
\end{promptbox}

\subsection{Shape-Shape Compositions}
\label{app:shape-shape}

To generate shape-shape compositions, we sample from the following pools of entities.

\begin{promptbox}[Shape-Shape Characteristics]
\small
\textbf{Objects:} backpack, amplifier, boar, toaster, vacuum, bonsai, amplifier, crystal, cactus, camera, chipmunk, refrigerator, toaster, compass, totem, telescope, dagger, projector, dolphin, office chair, radio, drone
\end{promptbox}

\begin{promptbox}[Generation Prompt]
\small
\texttt{"Generate an image of an animal/object that appears to be the logical composition of a \var{obj1} and a \var{obj2}. Make the composition/merging of the two fluid and tasteful, such that the outcome is one cohesive animal/object. Make the background white and do not add anything else. Use a realistic art-style."}
\end{promptbox}

\subsection{Fully Novel Entities}
\label{app:novel-entities}

Fully novel entities are generated from compositional design specifications spanning five attribute categories. The following are a representative subset of the characteristic and design specification lists used to compose each novel object prompt.

\begin{itemize}
    \item \textbf{Silhouettes:} ``lopsided hourglass with one chamber partially collapsed inward'', ``asymmetric clamshell that never fully closes'', ``dense knot-like mass with three lobes fused unevenly'', ``flattened sphere stretched diagonally as if pulled while soft'', ``blocky central volume pierced by an off-axis tunnel'', ``tall obelisk-like form warped into a gentle S-curve'', ``compact puck shape with one side bulging outward unnaturally'', ``clustered pebble-like forms fused into a single body'', ``torso-like volume missing its top and bottom planes'', ``squat pyramid whose faces bow inward instead of outward'', etc.

    \item \textbf{Materials:} ``slimy translucent green elastomer with suspended cloudy streaks'', ``pink fluffy synthetic fiber compacted into a rigid solid'', ``charred-looking polymer with a soft rubbery core'', ``oily black resin that reflects light unevenly'', ``milky silicone infused with darker fibrous strands'', ``ceramic glaze that appears cracked but is perfectly smooth'', ``carbon-fiber composite distorted into melted-looking waves'', ``semi-transparent plastic resembling congealed candy'', ``rubberized foam sealed under a glossy hard shell'', ``bioplastic with faint organic veining like fat or cartilage'', ``matte stone-like polymer that looks eroded but new'', ``frosted gelatinous material that appears wet but is solid'', etc.

    \item \textbf{Structural Rules:} ``exactly six curly protrusions that twist in alternating directions'', ``three hollow prongs that bend slightly toward a shared center'', ``one thick structural arm that dominates all other elements'', ``a ring-like element partially embedded and partially exposed'', ``five uneven spikes emerging only from concave regions'', ``a continuous internal void visible through irregular openings'', ``two mirrored appendages and one deliberately mismatched third'', ``structural elements that appear stacked but never align vertically'', ``four bulbous extensions connected by thin neck-like bridges'', ``a rigid outer frame constraining a visibly softer inner body'', etc.

    \item \textbf{Surface Details:} ``clusters of blunt micro-spikes that feel biological but artificial'', ``puckered dimples scattered unevenly like pinched clay'', ``fine wrinkles radiating outward from structural stress points'', ``patches of glossy smoothness interrupting an otherwise matte skin'', ``micro-ridges that abruptly stop and restart without pattern'', ``subtle surface sagging as if the material barely holds its shape'', ``tiny vent-like holes that suggest pressure release but do nothing'', ``polished seams that zigzag unpredictably across the object'', ``areas that appear stretched thin over an internal structure'', etc.

    \item \textbf{Palettes:} ``sickly pastel green with oily black shadows'', ``cotton-candy pink contrasted with industrial dark gray'', ``bone white smeared with muted bruise-purple undertones'', ``smoky translucent amber paired with dead matte charcoal'', ``pale fleshy beige offset by sharp graphite accents'', ``desaturated teal fading unevenly into off-black'', ``chalky lavender with dirty metallic silver edges'', ``warm nicotine-yellow contrasted with deep asphalt gray'', etc.
\end{itemize}

\begin{figure*}[!t]
    \centering
    \includegraphics[width=\linewidth]{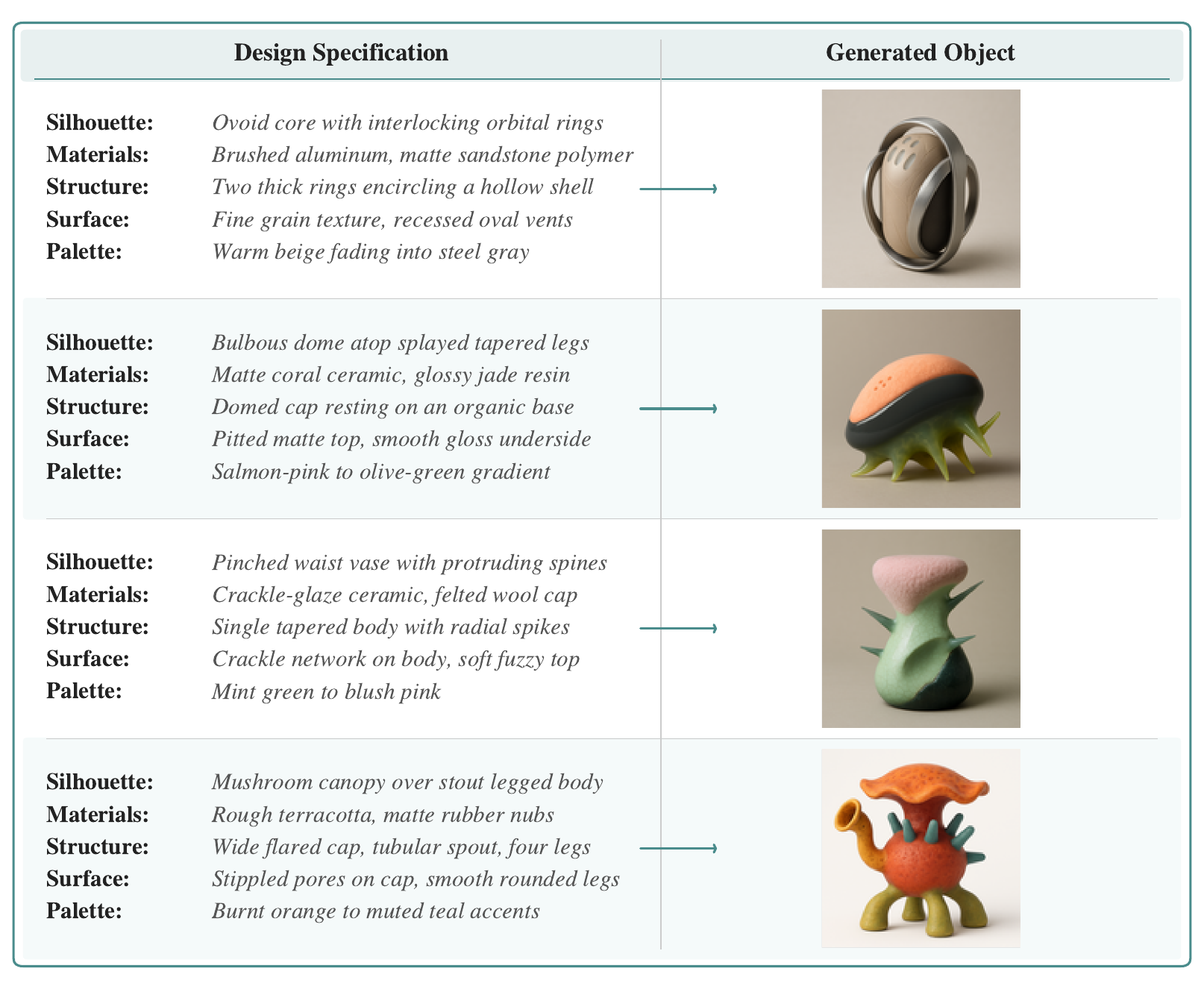}
    \caption{Example prompt compositions used to generate novel entities. Each row shows a unique design specification (left) and the resulting generated object (right).}
    \label{fig:novel_gen}
\end{figure*}

\Cref{fig:novel_gen} shows example prompt compositions used to generate novel entities from these design specifications.

\FloatBarrier
\newpage
\section{Image Perturbation Prompts and Settings}
\label{sec:gen-prompts}

To augment each of our objects with our variety of perturbation types, we use the templates, objects, and prompts listed and described below. Template variables are shown in \var{blue}. Perturbations fall into two broad groups: \emph{low-level edits} that modify surface properties without changing object structure, and \emph{higher-level edits} that alter the object's shape, composition, or semantic identity. Perturbations within each group are presented in the same order as in the main paper (\Cref{sec:perturbations}).

\subsection{Low-Level Edits}
\label{app:lowlevel-edits}

The following five perturbation types are applied \emph{programmatically} (i.e.\ without a generative model) and do not alter the object's shape or structure.

\subsubsection{Gaussian Noise}

At each level $\ell$, we add i.i.d.\ Gaussian noise $\epsilon \sim \mathcal{N}(0, \sigma_\ell^2 \mathbf{I})$ to the image, where $\sigma_\ell$ increases linearly with $\ell$. Because each level is applied to the output of the previous level (compounding), the cumulative noise intensity grows across the perturbation sequence.

\subsubsection{Scale}

We progressively down-sample the image using nearest-neighbor interpolation, reducing spatial resolution at each level. The scale factor decreases linearly from near-original resolution at level~1 to a highly reduced image at the final level, probing the spatial granularity at which models can still maintain concept mappings.

\subsubsection{Pixelation}

Similar to scale, we apply nearest-neighbor down-sampling followed by nearest-neighbor up-sampling back to the original resolution, producing a mosaic-like effect. The block size increases with each level, progressively removing fine-grained spatial detail while preserving the overall color distribution.

\subsubsection{JPEG Compression}

We apply JPEG compression with progressively decreasing quality factors across levels. This introduces color banding, block artifacts, and high-frequency information loss, allowing us to probe the effect of compression artifacts on concept judgments \citep{dodge2016understandingimagequalityaffects}.

\subsubsection{Color Shift}

We apply an arbitrary hue rotation of increasing intensity at each level. The hue shift angle increases linearly across levels, progressively altering the object's color palette while leaving shape and texture entirely intact.

\subsection{Higher-Level Edits}
\label{app:highlevel-edits}

The following six perturbation types involve generative editing or hybrid generative-programmatic pipelines. Unless otherwise noted, all generative perturbations are produced using Gemini-2.5 Flash Image and Gemini-3 Pro Image.

\subsubsection{Texture Shift}

We generate texture perturbations by transferring a target texture (e.g., a slime-like surface) onto an object (e.g., a golden retriever), while largely preserving its global shape and structure. We then linearly interpolate between the original and textured images across 20 levels: $x_\ell = (1 - t_\ell)\, x_{\text{orig}} + t_\ell\, x_{\text{textured}}$, where $t_\ell = \ell / L$ for $\ell \in \{1, \ldots, L\}$ with $L = 20$, such that the visible texture intensity increases smoothly. This perturbation involves a hybrid generation process combining both generative and programmatic augmentations.

\subsubsection{Background Replacement}

Background perturbation involves two sub-steps: we first generate a set of target scene images, and then we programmatically alpha-blend the background at each level, compositing the original object on top.

\begin{promptbox}[Background Scene Generation Prompt]
\small
Generate a photorealistic photograph of \var{scene}. Show the full scene
filling the entire frame, photographed from eye level. No people, animals,
or prominent foreground objects --- just the environment/setting itself.
Detailed, well-lit, natural-looking photograph.
\end{promptbox}

We use the following background scenes for the perturbations:
\begin{quote}\small\itshape
``a wizard's study with bookshelves, scrolls, candles, and arcane instruments'',
``a cozy English pub with wooden beams, bar stools, pint glasses, and dartboard'',
``a 1950s American diner with red vinyl booths, a jukebox, and checkered floor'',
``a Japanese zen garden with raked sand, smooth stones, a bamboo fountain, and bonsai'',
``a space station interior with control panels, round windows showing stars, and metal walls'',
\ldots~(25 scenes total)
\end{quote}

We compute the background blend as follows:
\begin{align*}
  I_k = \alpha\text{-blend}&\!\left(\text{bg\_color},\; I_{\text{target}},\; \tfrac{k}{N}\right) \\
       &\oplus\; \text{composite}(I_{\text{original}})
\end{align*}

\subsubsection{Style Degradation}

To generate style degradation perturbations, we use a fixed 20-step trajectory ranging from a photorealistic art style to a largely blank canvas. We describe the prompt as well as the fixed trajectory in \Cref{tab:style-trajectory} below.

\begin{promptbox}[Style Degradation Prompt]
\small
Slightly reduce the artistic quality and detail of this image. The result
should look like \var{trajectory\_description}. Make only a small change from
the current image --- keep the same general pose and composition.
\end{promptbox}

\begin{table}[!t]
\centering
\footnotesize
\setlength{\tabcolsep}{4pt}
\renewcommand{\arraystretch}{1.1}
\begin{tabular}{@{}cp{0.78\columnwidth}@{}}
\toprule
\textbf{Step} & \textbf{Target Description} \\
\midrule
1  & High-quality digital painting; smoother shading and softer edges than a photograph \\
2  & Professional digital illustration; smoother color transitions, slightly less realistic lighting \\
3  & Polished illustration with subtly flatter shading; most detail preserved \\
4  & Clean illustration with moderately flat shading \\
5  & Illustration with noticeably flat colors and slightly imprecise line work \\
6  & Simple, clean illustration with flat unblended colors and basic outlines \\
7  & Simplified graphic with about 10--15 flat color regions, no gradients \\
8  & Bold, blocky graphic using only about 6--8 large flat color regions \\
9  & Rough graphic with only 3--4 flat colors and thick uneven outlines \\
10 & Crude rendering using 2--3 colors in roughly drawn overlapping shapes \\
11 & Very rough single-color sketch; a few thick imprecise strokes \\
12 & A few thick, sloppy strokes of a single dark color loosely hinting at the outline \\
13 & 3--4 rough overlapping dark strokes that very vaguely suggest the original mass \\
14 & 2--3 thick, rough smears of dark paint in approximately the original shape \\
15 & A single thick rough brushstroke or blob of dark paint \\
16 & An amorphous dark smudge; a rough mass of dark color \\
17 & A faint, irregular dark stain or wash with almost no visual information \\
18 & A barely visible, faded smudge of muted color \\
19 & A nearly blank image with only the faintest discoloration \\
20 & An essentially blank white image \\
\bottomrule
\end{tabular}
\caption{Style degradation trajectory (20 steps).}
\label{tab:style-trajectory}
\end{table}

\subsubsection{Shape Deformation}

To deform the shape of each object, we use the following prompt.

\begin{promptbox}[Shape Deformation Prompt]
\small
For the following image, complete this visual edit:
Strongly deform, warp, or mutate the overall shape and silhouette of the
subject. The change should be bold and unmistakable --- significantly distort
proportions, twist or melt parts of the body, or fracture and reshape the
outline. Keep the rest of the image unchanged.
\end{promptbox}

\subsubsection{Part Addition}

To add a new part to each object at each new level $\ell$, we simply prompt the generative model as described below.

\begin{promptbox}[Part Addition Prompt]
\small
For the following image, complete this visual edit:
Add a large, clearly visible extra part, limb, or appendage to the subject.
Make it bold --- not a subtle bump, but a prominent new structure that
obviously changes the subject's form. Keep the rest of the image unchanged.
\end{promptbox}

\subsubsection{Part Removal}

To produce part removal perturbations, we first generate a list of ``removable'' parts for each object using GPT-4o Mini. For a golden retriever, for instance, this list would include its paws and legs, its ears, nose, and eyes, its tail, and finally, its torso. At each next level $\ell$, the $\ell$-th part from the list is targeted for removal; thus, we structure each list to prioritize conducting more semantically and visually significant changes last.

\begin{promptbox}[Part List Generation Prompt]
\small
Analyze this image of a(n) \var{obj} carefully.

List exactly \var{n\_parts} distinct, removable parts of this specific subject,
ordered from MOST visually prominent/obvious to LEAST obvious.

Rules:
\begin{itemize}[nosep, leftmargin=1.5em]
  \item Each part must be a specific, concrete body part or appendage
        (e.g.\ ``tail'', ``left front leg'', ``right antenna'', ``dorsal fin'')
        --- NOT abstract concepts like ``texture'' or ``color''.
  \item Parts should be things that could realistically be erased/removed from
        the image.
  \item Be specific about LEFT vs RIGHT, FRONT vs BACK when applicable.
  \item Include both large parts (legs, wings, head) and small parts
        (individual toes, whiskers, claws).
  \item If the subject has fewer than \var{n\_parts} truly distinct parts,
        repeat removal of the same type of part but specify differently
        (e.g.\ ``front left leg'' then ``front right leg'').
  \item For later entries when obvious parts are exhausted, include things like
        ``left eye'', ``nose'', ``mouth'', or describe portions of the body
        (``upper torso'', ``lower abdomen'').
\end{itemize}
\end{promptbox}

\begin{promptbox}[Part Removal Prompt]
\small
For the following image, complete this visual edit:
Remove the \var{part\_k} from the subject in the image. Completely erase it so
there is a clear, visible gap or absence where the \var{part\_k} used to be.
Keep everything else unchanged.

Parts already removed in previous levels: [\var{parts\_list}]. You MUST remove
a DIFFERENT part that has NOT already been removed.

Keep the rest of the image unchanged.
\end{promptbox}

\FloatBarrier
\newpage
\section{Dataset Generation, Validation, and Quality Control Details}
\label{sec:generation}

Perturbations requiring generative editing are created using both Gemini-2.5 Flash and Gemini-3 Pro with a dynamic prompt including the source image \citep{comanici2025gemini25}. To validate that each perturbation is sufficient for experimentation, we employ a two-stage quality control pipeline using Gemini-2.5 Flash as a VLM judge. All judge prompts share the following preamble, which contextualizes the task for the VLM:

\begin{promptbox}[Project Context Preamble]
\small
You are part of a research pipeline studying learning biases in Vision-Language Models (VLMs). We are generating datasets of progressively perturbed images to measure how sensitive VLMs are to different visual properties (shape, texture, color, scale, etc.).

For each object, we generate up to 10 levels of compounding perturbations along a single axis (e.g.\ 10 levels of shape deformation). Each level should be MORE visually/semantically different from the original than the previous level --- this creates a monotonically increasing difficulty curve for VLM recognition.

The KEY REQUIREMENT is that each perturbation level produces a CLEAR, VISIBLE change along the intended axis ONLY. Changes to other axes (e.g., shape changing when we only asked for texture) confound the experiment. The perturbation should be as pure and isolated as possible.

At high levels, the object may become unrecognizable --- this is expected and desired. The goal is to find the point at which VLMs can no longer identify the object.
\end{promptbox}

\subsection{Per-Level Judge}
\label{app:per-level-judge}

At each level $\ell$, the judge receives both the source image $x_{\ell-1}$ and the perturbed output $\hat{x}_\ell$ and evaluates whether (i) the model's edit was ``clearly and visibly applied'' and (ii) the model didn't introduce unwanted changes along other perturbation axes (e.g., changing the shape when only color was requested), as such changes could contaminate our axis-specific analysis. When the judge rejects a generation, it provides a revised instruction with details specific to the particular object depicted in the image, as the generative model is initially prompted with a generic prompt that isn't tailored to the specific object, and then it prompts the generative model to re-attempt the perturbation up to 10 times. For saturable perturbation types (scale, part removal, and artistic style) the judge additionally assesses whether the perturbation has reached a natural limit where it can no longer be applied (e.g., the object is too small to shrink further, no parts remain to remove). When saturation is detected, the sequence terminates at the current level. We use the following prompt:

\begin{promptbox}[Per-Level Judge Prompt]
\small
[Project Context Preamble]

You are evaluating a single step in this pipeline. The original image (first) shows a(n) \var{obj}.
The perturbation axis is: ``\var{p\_type}''
The specific edit instruction was: ``\var{perturbation\_desc}''

Evaluate against these criteria:
\begin{enumerate}[nosep, leftmargin=1.5em]
  \item Is the requested edit (``\var{p\_type}'') clearly and visibly applied in the new generation?
  \item Does the new generation show a visible change compared to the previous level?
  \item Are there unwanted changes along OTHER axes? (e.g.\ shape changing when only texture was requested)
        --- Changes along the requested axis are always welcome, even if dramatic.
        --- Changes along other axes confound the experiment and should be flagged.
\end{enumerate}

If the edit failed, write a revised\_prompt that is far more specific to the actual subject you see in the first image.
--- Look at the image carefully and describe the subject's specific parts, colors, textures, or spatial layout.
--- Instead of generic instructions, describe exactly what to change and where on this particular subject.
--- The revised prompt should be a complete, self-contained instruction for the image editor.

Respond ONLY in valid JSON with these fields:
\begin{verbatim}
{
  "passed": true or false,
  "reason": "one or two sentences explaining why",
  "revised_prompt": "if failed, complete object-
  specific rewrite; if passed, null",
  "saturated": true or false (saturable type only)
}
\end{verbatim}

Only set passed=true if the requested edit is clearly visible. The \var{obj} does NOT need to remain recognizable --- progressive degradation is expected and desired.
\end{promptbox}

\subsection{Global Sequence Judge}
\label{app:global-sequence-judge}

After the full sequence is generated, a global judge receives the complete sequence $(x_0, x_1, \dots, x_\ell)$ and evaluates whether visual distance from $x_0$ increases smoothly across all levels and the specific perturbation axis. The judge identifies \emph{undesirable levels}, or those that regress back toward the original or stagnate for three or more consecutive steps, and returns their indices. For each undesirable level, the corresponding image is regenerated from its predecessor using the same per-level judge protocol as before, and the sequence is re-evaluated, for up to three global rounds. As an additional safeguard, the global judge is also used at intervals of every five levels during generation, allowing mid-sequence corrections before the full sequence is complete. We use the following prompt:

\begin{promptbox}[Global Sequence Judge Prompt]
\small
[Project Context Preamble]

You are evaluating a full sequence of \var{n} progressively perturbed images of a(n) \var{obj}.
The perturbation axis is: ``\var{p\_type}''

The first image is the ORIGINAL (unperturbed). The following \var{n} images are levels 1 through \var{n}, each generated from the previous level.

The sequence should exhibit MONOTONIC PROGRESSIVE DIVERGENCE from the original along the ``\var{p\_type}'' axis:
\begin{itemize}[nosep, leftmargin=1.5em]
  \item Each level should look MORE different from the original than the previous level.
  \item The visual distance from the original should strictly increase (or at least not decrease).
  \item There should be no ``resets'' where a later level suddenly looks more like the original.
  \item There should be no long plateaus where multiple consecutive levels look identical.
\end{itemize}

For each level, assess whether it maintains the monotonic progression.

Respond ONLY in valid JSON:
\begin{verbatim}
{
  "monotonic": true or false,
  "bad_levels": [list of 1-indexed level numbers
                  that break monotonicity],
  "reasoning": "brief overall assessment"
}
\end{verbatim}

A level is ``bad'' if:
\begin{enumerate}[nosep, leftmargin=1.5em]
  \item It looks MORE similar to the original than the PREVIOUS level (regression), OR
  \item It looks virtually identical to the previous level (stagnation in a run of 3+ stagnant levels).
\end{enumerate}

Be strict about regressions but lenient about minor plateaus (2 similar levels are OK; 3+ are not).
\end{promptbox}

\subsection{Post-Hoc Sequence Cleaning}
\label{app:post-hoc-cleaning}

These judges alone are not sufficient to ensure a clean degradation in the target object along the specified perturbation axes, thus, we also conduct post-hoc sequence cleaning using Gemini-2.5 Flash. We prompt the model to score each level on a 0--100 scale indicating how intact the object remains, allowing us to identify levels which appear to be duplicates of other levels, or which linger outside of acceptable visual degradation at its particular level, thus stripping our perturbation sequences of redundant levels and images.

\begin{promptbox}[Sequence Scoring Prompt]
\small
You are evaluating a sequence of images where a(n) \var{obj} is progressively perturbed.
The FIRST image is the ORIGINAL. The next \var{n} images are levels 1--\var{n}.

For EACH level, rate how intact/complete the \var{obj} still is on a scale of 100 (100 = fully intact like original, 0 = completely removed/gone). Focus on how much of the object's structure remains visible.

Respond ONLY in valid JSON: \texttt{\{"scores": [s1, s2, ..., s\var{n}]\}}
\end{promptbox}

For cases where this cleaning reduces the number of levels for a certain perturbation below ten total levels, we identify the largest score gaps in the remaining sequence and generate intermediate levels to fill them, targeting a score midway between adjacent levels. We also re-generate relevant perturbation levels using Gemini-3 Pro Image. Finally, we, the authors, manually validate and curate the dataset prior to conducting both model and human experimentation.

\subsection{Manual Author Validation}
\label{sec:app-author-val}

Manual author validation was also conducted to validate the cleanliness of the dataset. 100 image pairs were sampled across all object categories and perturbation levels and the high-level perturbation types, such that, if the first image in the pair is of object $X$ with perturbation type $Y$ at level $Z$, then the second image in the pair will be of object $X$ with perturbation type $Y$, but at level $Z+1$. We then manually confirm if $Y$ was correctly applied to $X$ to logically yield the following image in the trajectory, finding that, of the 100 image pairs, 18 of them were undesirable or noisy, that is, the perturbation wasn't properly applied, or it was properly applied, but some other aspect of the object was also changed inappropriately (e.g. removing a certain part for the part removal perturbation, but also adding another part elsewhere on the object).

\newpage

\FloatBarrier
\section{Experimental Details}
\label{sec:experimental-details}

This section provides full prompting details for each of the three experimental paradigms described in \Cref{sec:experiments}. All experiments are conducted under greedy decoding unless otherwise noted.

\subsection{Name Generation from Multi-Image In-Context Learning}
\label{app:name-gen}

We iterate over each image in our data sample, as well as all of its available perturbation types and levels, where the main (non-perturbed) visual stimulus is paired with its nonce word caption using one of the following five caption templates:

\begin{itemize}[itemsep=2pt]
    \item ``This image is best described by the reference: \textcolor{blue}{nonce label}'',
    \item ``This image shows \textcolor{blue}{nonce label}'',
    \item ``In this image, we see \textcolor{blue}{nonce label}'',
    \item ``This image depicts \textcolor{blue}{nonce label}'',
    \item ``The subject of this image is \textcolor{blue}{nonce label}''.
\end{itemize}

After the in-context pool (containing the target image-caption pair and four visually similar distractors from PixMoCap; see Section~4.1 of the main paper), the perturbed stimulus is presented last with the following fill-in-the-blank prompt:

\begin{promptbox}[Fill-in-the-Blank Generation Prompt]
\small
This image is best described by the reference: \_\_\_\_
\end{promptbox}

All models use greedy decoding and generate responses to fill the blank, re-generating up to three times if the response is shorter than 2 characters. We construct nonce words by prompting GPT-4o to generate candidates and filtering for nonce words with exactly three tokens in length.

\subsection{Token Probabilities Given Multi-Image In-Context Learning}
\label{app:tokprob}

Using an identical setup to the multi-image generation, we present the shuffled image pools to the models, but rather than a fill-in-the-blank task, we provide the final image caption with the target nonce reference and compute the reference probability using:
\[
  \frac{1}{N}\bigl(\log P(r \mid \mathcal{C})\bigr) = \frac{1}{N}\Bigl(\sum_{i=1}^{N} \log P(t_i \mid t_{<i}, \mathcal{C})\Bigr)
\]
where $r$ is the target nonce reference, $N$ is the token-length of the reference, $t_1, t_2, \ldots, t_N$ are its constituent tokens, and $\mathcal{C}$ is the full in-context image pool with captions, the instruction, and the final target caption. We compute $\log P(t_i \mid t_{<i}, \mathcal{C})$ by applying log-softmax over the model's output logits and selecting the entry for~$t_i$. We also compute the probability of ``vanilla'' references (e.g.\ ``tree frog'' instead of the assigned nonce label for an image of a tree frog), to compare whether our models are genuinely acquiring the novel mappings or defaulting to labeling using familiar concepts. This task is only available for open-source models.

\subsection{Dual-Image Likert-Scale Rating}
\label{app:likert}

In this setup, the model is first shown the original image captioned ``Let's call the object in this image `\textcolor{blue}{[nonce word]}'.'' Then, it is shown the perturbed variant and asked to rate agreement with the statement ``Could both of these images be called `\textcolor{blue}{[nonce word]}'?'' on a scale from 1 to 7, where 1 = Strongly Disagree and 7 = Strongly Agree. The model responds with a single integer which we parse and collect. This setup aligns closely with the experimental setup used for our human study (see App.~\ref{sec:human-study}).

\newpage

\FloatBarrier
\section{Human Study Details}
\label{sec:human-study}

We conduct a crowd-sourced study through Prolific, collecting judgments from 30 anonymous native English speakers residing in the United States, Canada, UK, and Ireland on 800 unique image pair trials.

Inter-rater reliability, measured by the intra-class correlation coefficient ICC(2,$k$) which estimates the consistency of the \emph{mean} rating across $k=3$ raters per item, was $0.80$. On average, participants spent 6 minutes and 47 seconds on the full task. The task itself involved 80 image pair trials in addition to 3 instruction slides and 5 attention checks presenting participants with 2 unrelated images, flagging for those who don't respond with either ``Disagree'' or ``Strongly Disagree''; no participant failed any attention check. Study participants were paid on average \pounds 17.79/hour, and were informed of how their responses would be used for the purposes of the study, as well as their rights over their submitted data. Responses from participants who failed attention checks interspersed within the study were excluded from further analysis.

\begin{figure*}[!t]
    \centering
    \includegraphics[width=\linewidth]{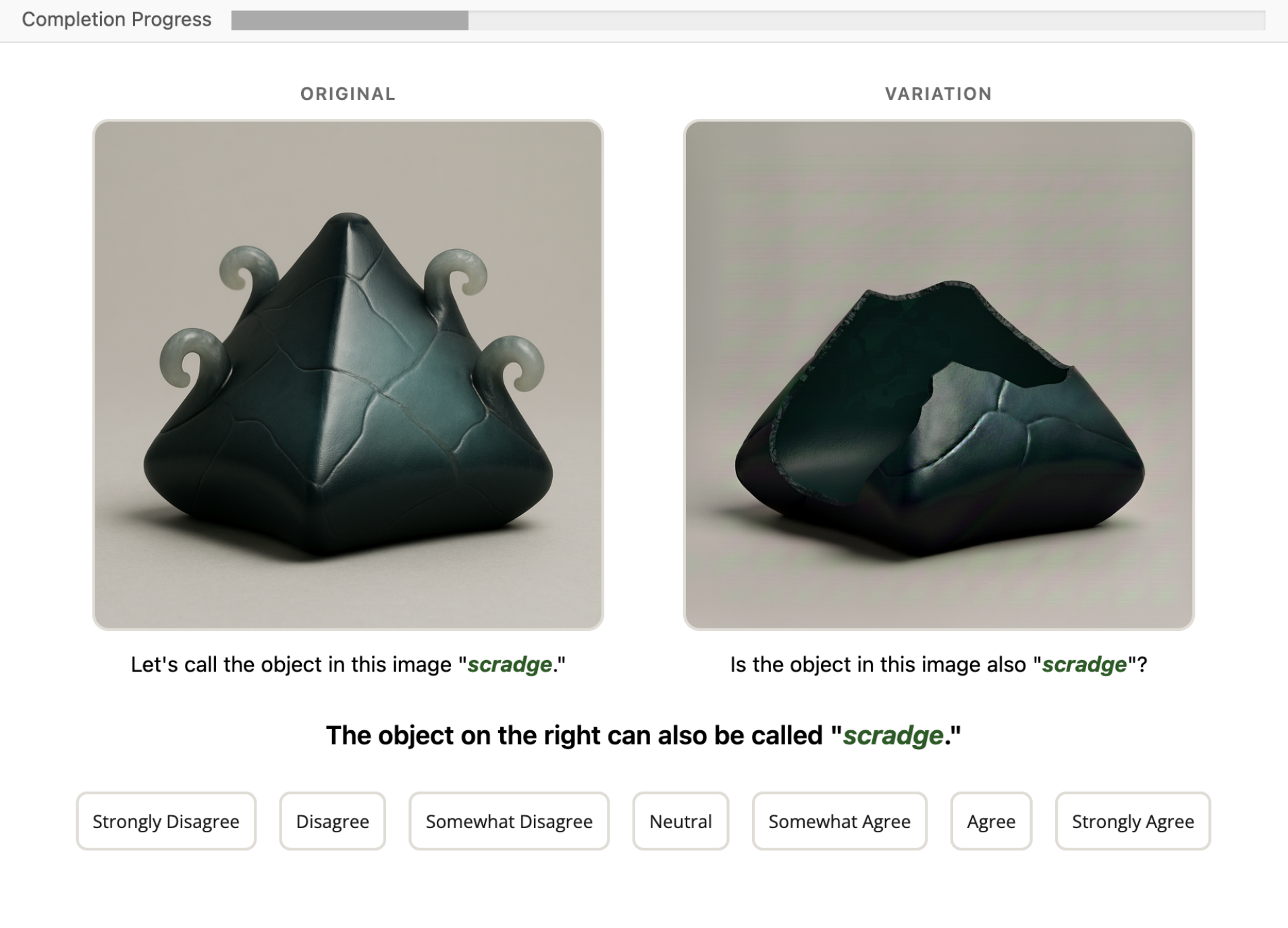}
    \caption{Example trial human participants observed during our study. Participants see the original image (left) and a perturbed variant (right), along with the nonce word, and respond on a 7-point Likert scale from ``Strongly Disagree'' to ``Strongly Agree.''}
    \label{fig:study_ex}
\end{figure*}

We show an example image pair trial with the user interface and setup a participant would have completed the study with in \Cref{fig:study_ex}.

\begin{figure*}[!t]
    \centering
    \includegraphics[width=\textwidth,height=0.85\textheight,keepaspectratio]{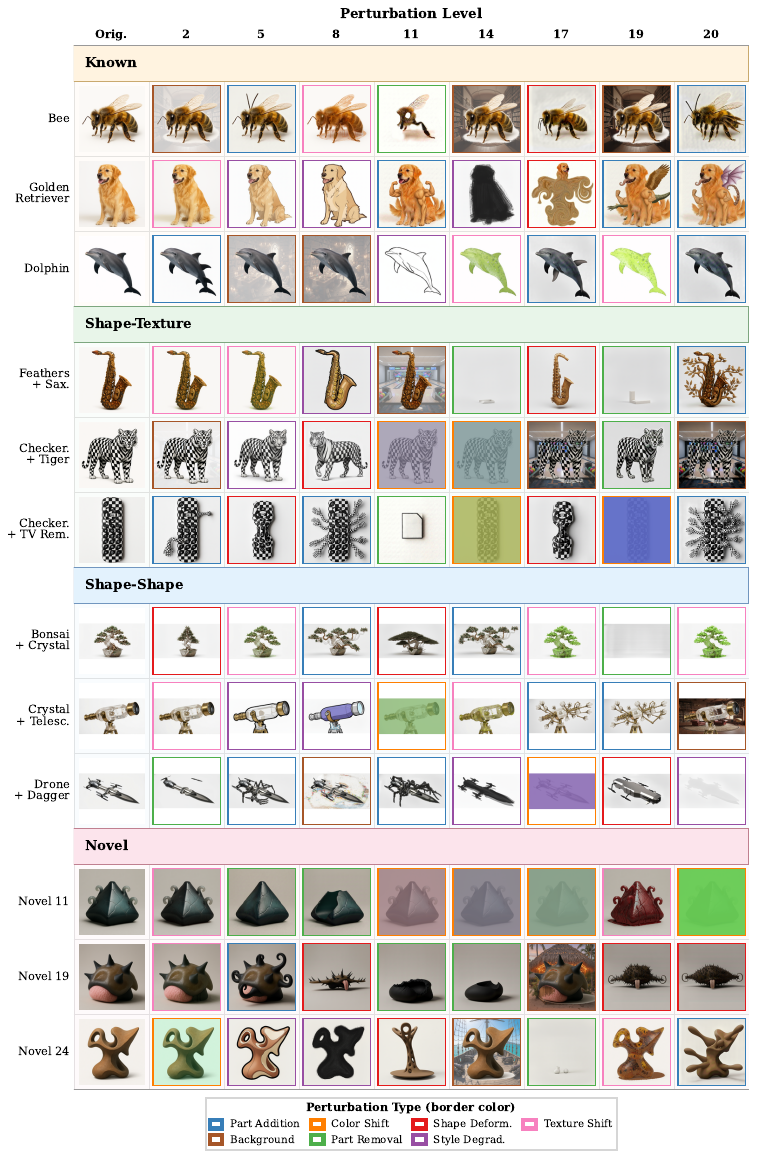}
    \caption{Sample of objects and perturbations from NVRD across the four object categories (Known, Shape-Texture, Shape-Shape, Novel) and eight perturbation levels examined in our human study. Each image border is color-coded by perturbation type (see legend). The leftmost column shows the original (unperturbed) base image for each object.}
    \label{fig:study-sample-grid}
\end{figure*}

\Cref{fig:study-sample-grid} presents the full sample of objects and perturbations used in our human study across all four object categories and eight perturbation levels.

\FloatBarrier
\newpage{Additional Results}
\label{sec:additional-results}

This section presents supplementary results organized by experimental paradigm.

\subsection{Name Generation Results}
\label{app:name-gen-results}

\begin{figure*}[!t]
    \centering
    \includegraphics[width=\textwidth]{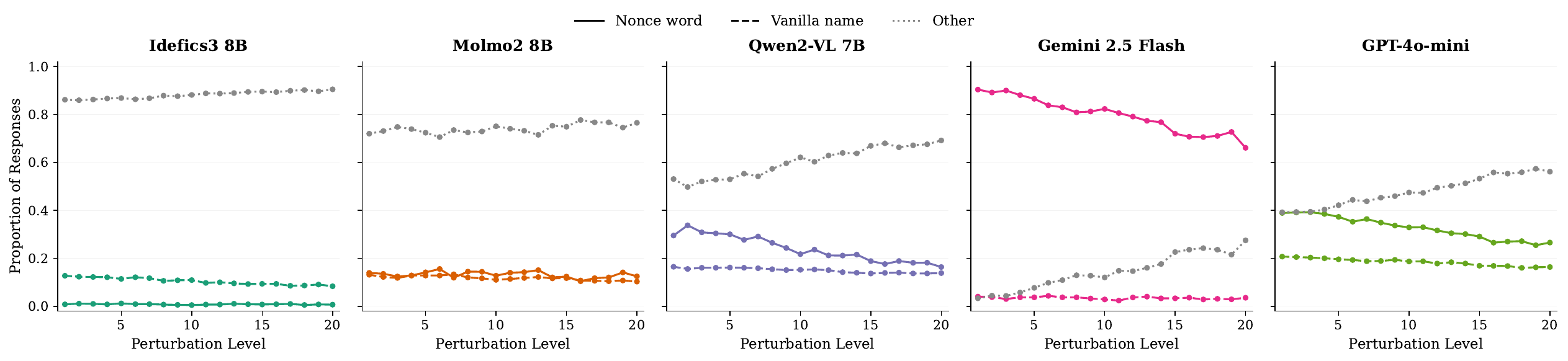}
    \caption{Nonce vs.\ vanilla label responses across models and perturbation levels.}
    \label{fig:nonce-vanilla-lineplot}
\end{figure*}

\begin{figure*}[!t]
    \centering
    \includegraphics[width=\textwidth]{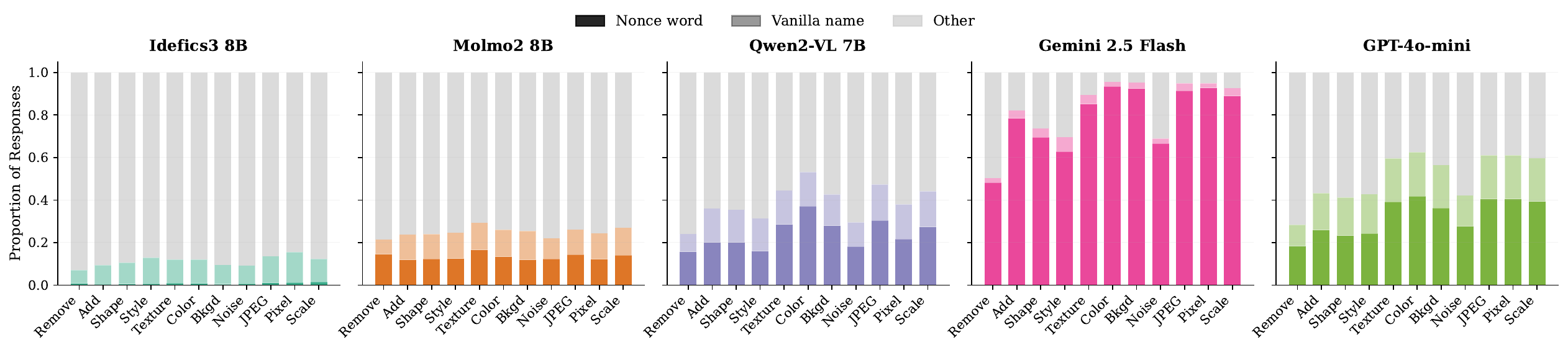}
    \caption{Nonce vs.\ vanilla label responses across models and perturbation types.}
    \label{fig:nonce-vanilla-pert-type}
\end{figure*}

\Cref{fig:nonce-vanilla-lineplot,fig:nonce-vanilla-pert-type} show the breakdown of nonce vs.\ vanilla label responses across perturbation levels and types, respectively, complementing the aggregate view in \Cref{fig:nonce-vanilla-obj-cat}.

\begin{figure*}[!t]
    \centering
    \includegraphics[width=\textwidth]{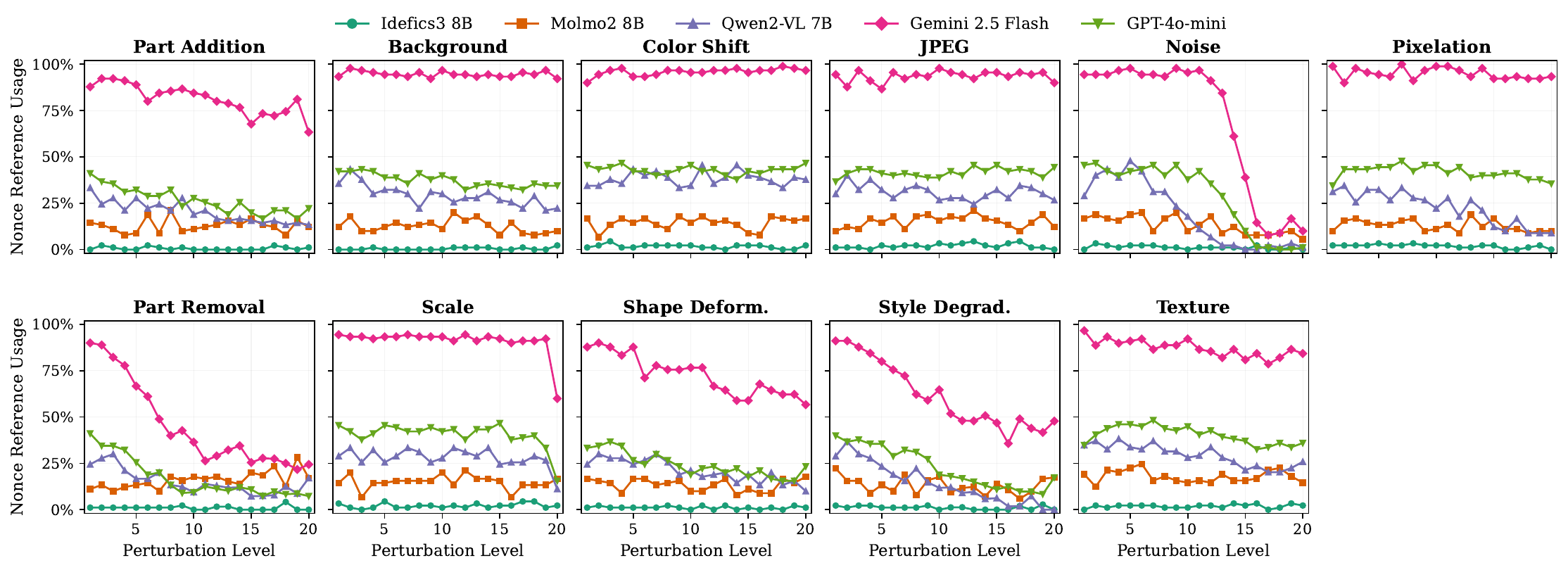}
    \caption{Model nonce reference usage across perturbation types and levels.}
    \label{fig:ref-gen-pert-type}
\end{figure*}

\Cref{fig:ref-gen-pert-type} provides the per-perturbation-type breakdown of nonce reference usage, showing how generation accuracy varies across all 11 perturbation axes.

\FloatBarrier
\subsection{Log Probability Results}
\label{app:logprob-results}

\begin{figure*}[!t]
    \centering
    \includegraphics[width=\textwidth]{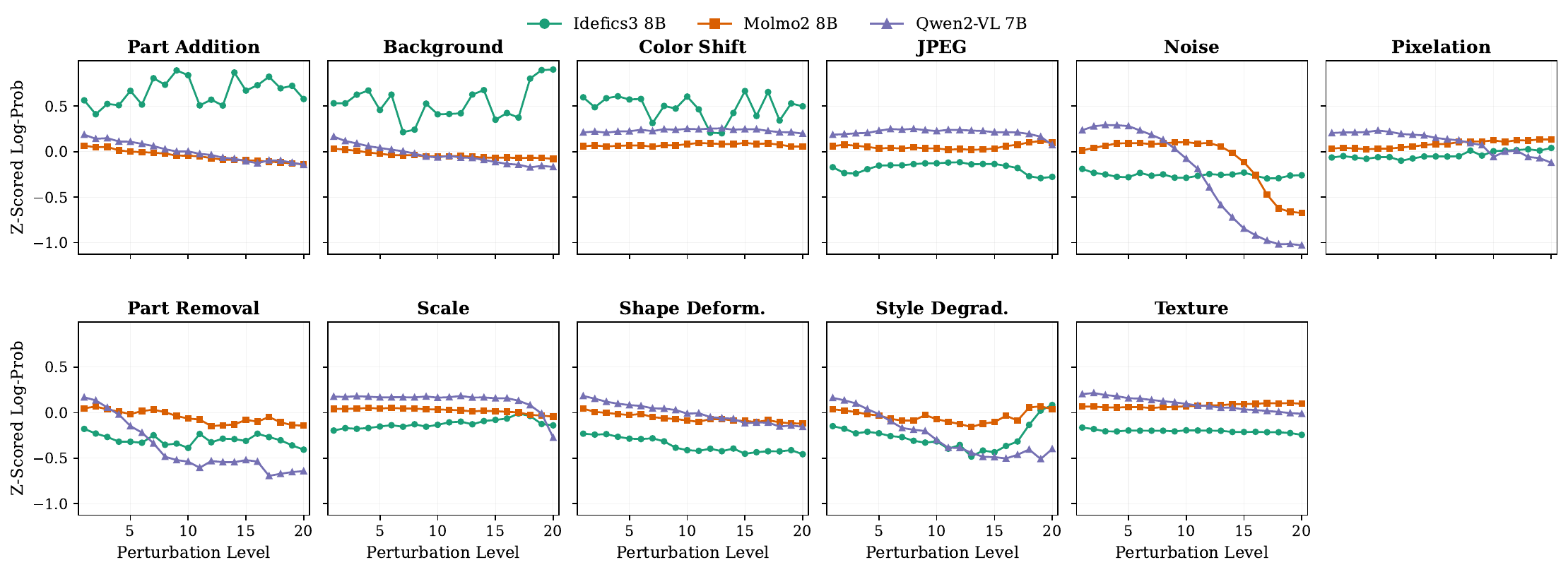}
    \caption{Model nonce $z$-scored log probabilities across perturbation types and levels.}
    \label{fig:ref-prob-pert-type}
\end{figure*}

\begin{figure*}[!t]
    \centering
    \includegraphics[width=\textwidth]{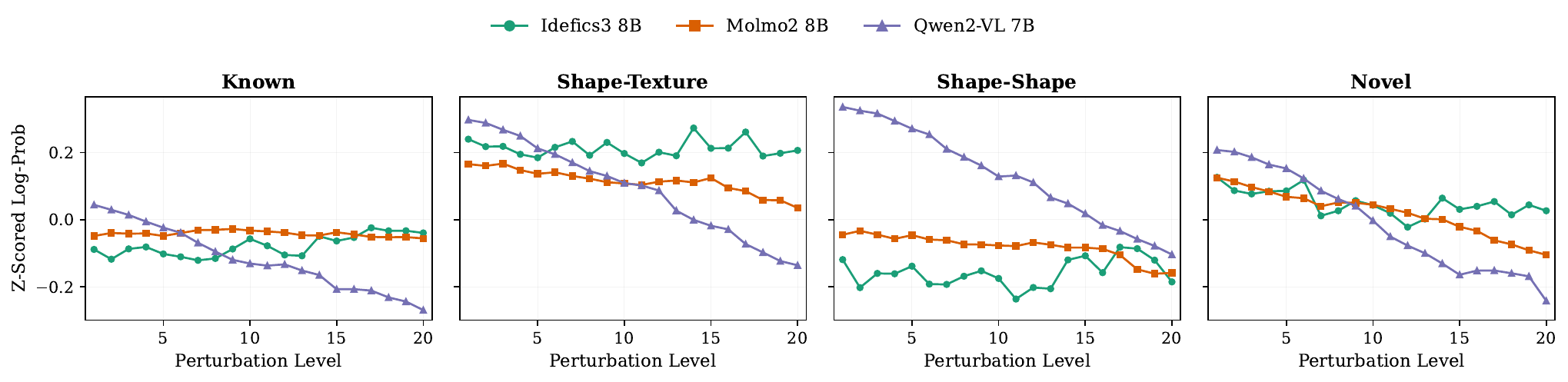}
    \caption{Model nonce reference $z$-scored log probabilities across object categories and perturbation levels.}
    \label{fig:ref-prob-obj-cat}
\end{figure*}

\Cref{fig:ref-prob-pert-type,fig:ref-prob-obj-cat} present the z-scored log probability analysis broken down by perturbation type and object category, respectively.

\FloatBarrier
\subsection{Likert Rating Results}
\label{app:g3}
\begin{figure*}[!t]
    \centering
    \includegraphics[width=\textwidth]{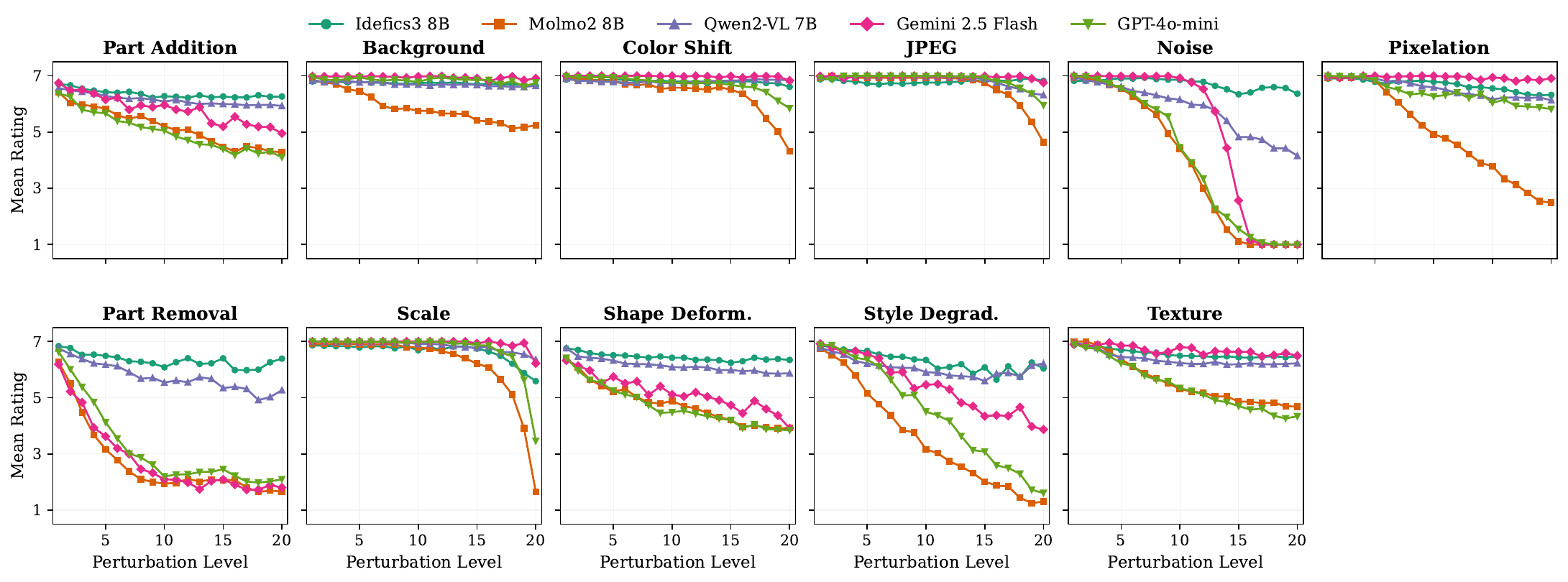}
    \caption{Model ratings across perturbation types and levels.}
    \label{fig:model-ratings-pert-type}
\end{figure*}

\begin{figure*}[!t]
    \centering
    \includegraphics[width=\textwidth]{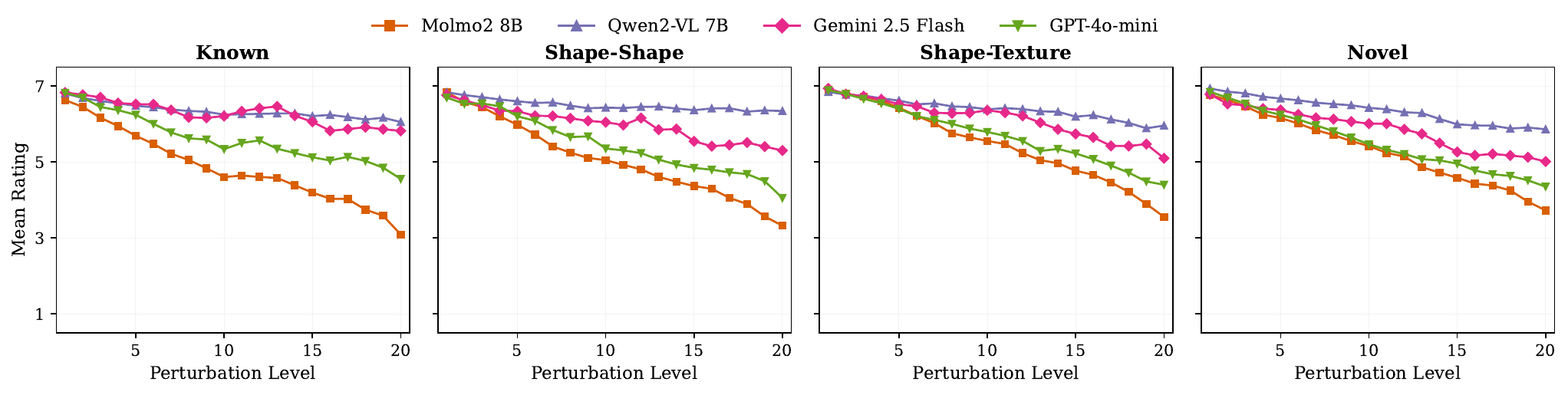}
    \caption{Model ratings across object categories and perturbation levels.}
    \label{fig:model-ratings-obj-cat}
\end{figure*}

\Cref{fig:model-ratings-pert-type,fig:model-ratings-obj-cat} show model Likert-scale ratings broken down by perturbation type and object category, respectively.

\FloatBarrier
\subsection{Human--Model Comparisons}
\label{sec:app-human-model-comparison}
\begin{figure*}[!t]
    \centering
    \includegraphics[width=\textwidth]{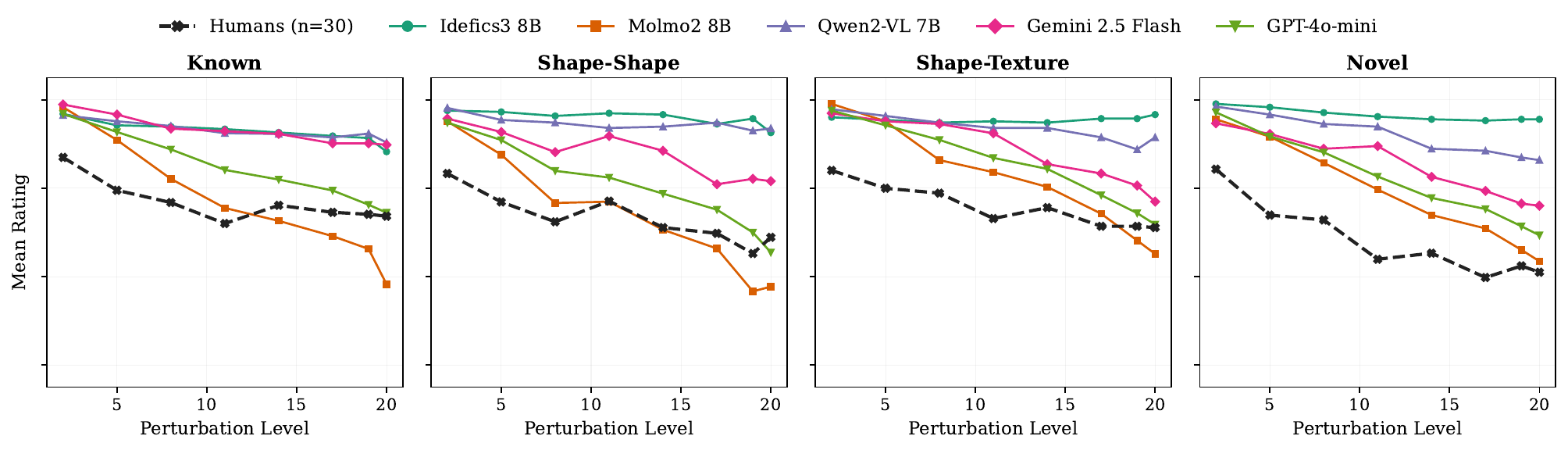}
    \caption{Human--model rating comparison across object categories and perturbation types.}
    \label{fig:human-model-obj-cat}
\end{figure*}

\begin{figure*}[!t]
    \centering
    \includegraphics[width=\textwidth]{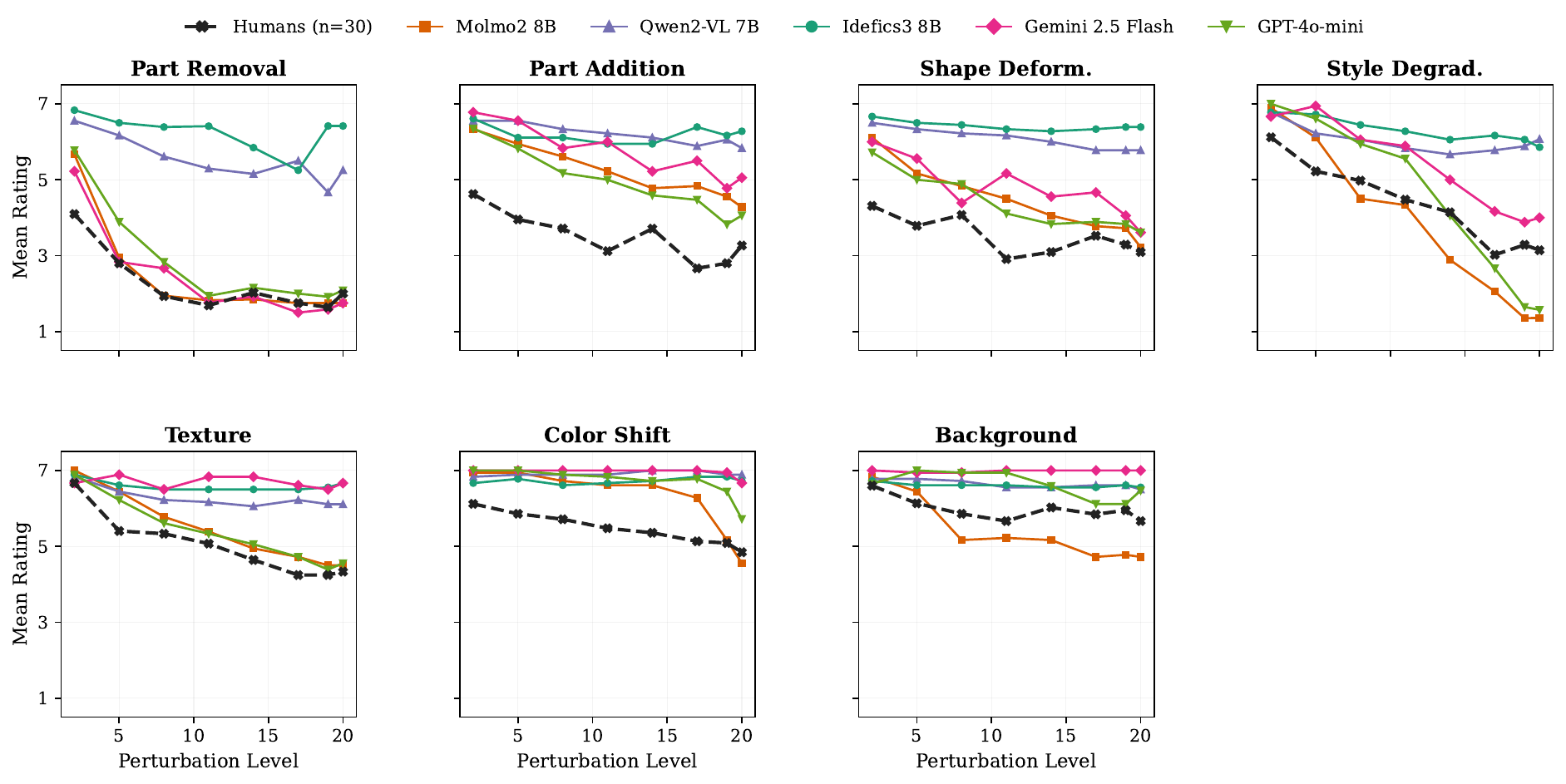}
    \caption{Human--model rating comparison across perturbation types.}
    \label{fig:human-model-pert-type}
\end{figure*}

\begin{figure*}[!t]
    \centering
    \includegraphics[width=\textwidth]{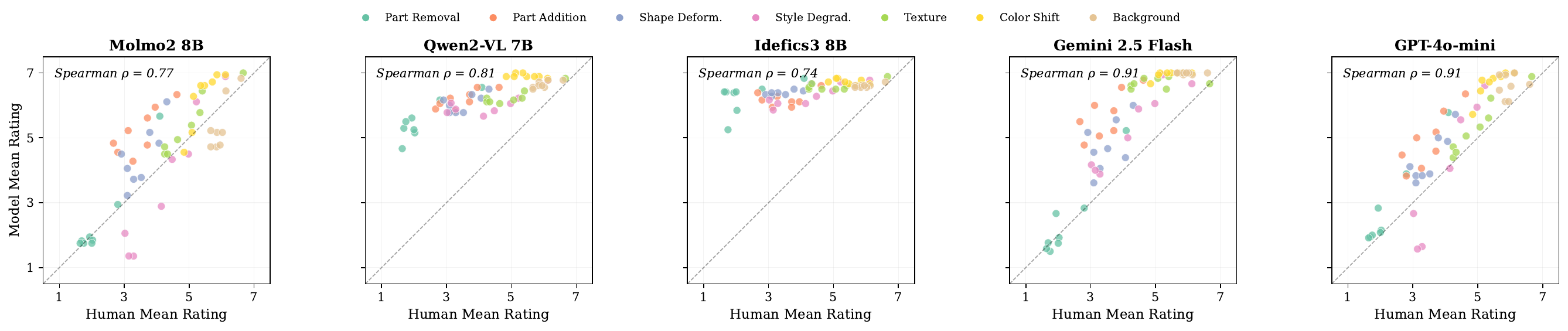}
    \caption{Scatterplot of human vs.\ model mean ratings across perturbation types and levels.}
    \label{fig:human-model-scatter-pert-type}
\end{figure*}

\begin{figure*}[!t]
    \centering
    \includegraphics[width=\textwidth]{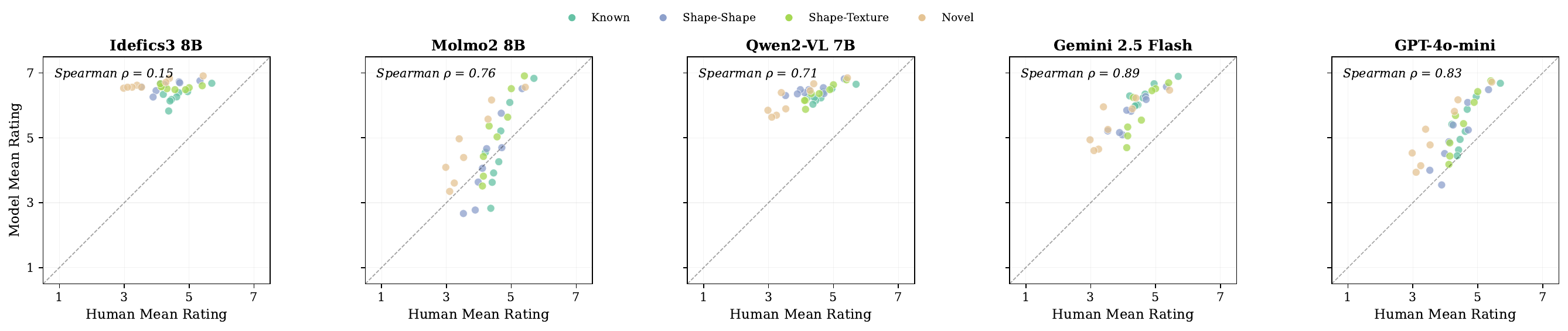}
    \caption{Scatterplot of human vs.\ model mean ratings across object categories and perturbation types.}
    \label{fig:human-model-scatter-obj-cat}
\end{figure*}

\begin{figure*}[!t]
    \centering
    \includegraphics[width=\textwidth]{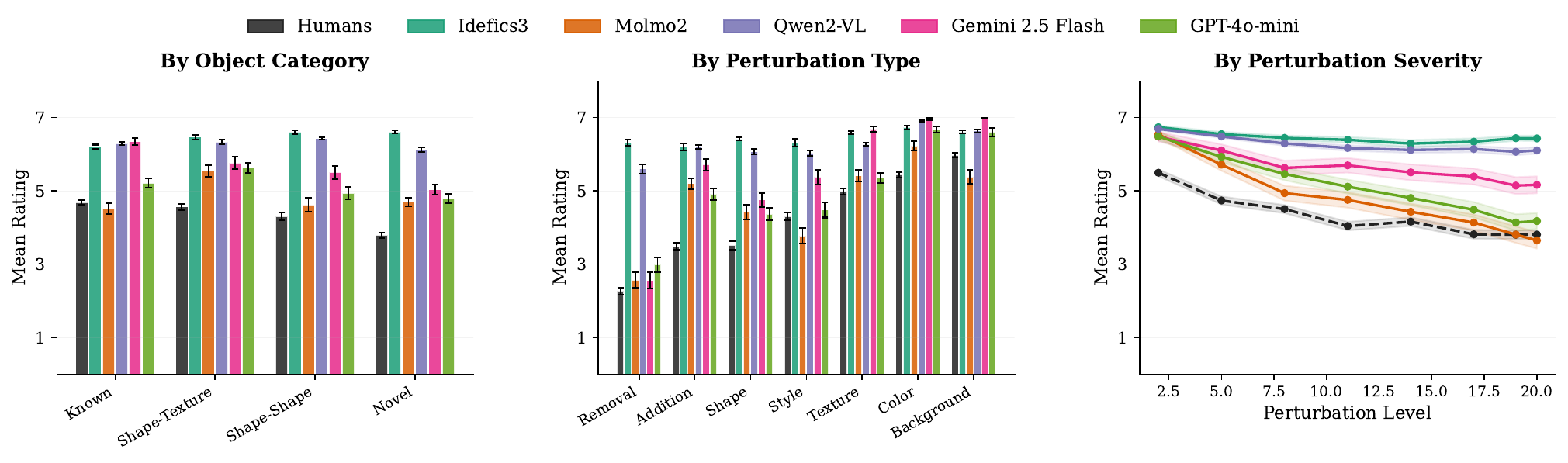}
    \caption{Human--model rating bar plot comparison across object categories, perturbation types, and perturbation severity.}
    \label{fig:human-model-barplot}
\end{figure*}

\Cref{fig:human-model-obj-cat,fig:human-model-barplot} compare human and model behavior across multiple views of the data. \Cref{fig:human-model-obj-cat} breaks down the comparison by object category, \Cref{fig:human-model-pert-type} by perturbation type, \Cref{fig:human-model-scatter-pert-type,fig:human-model-scatter-obj-cat} show scatterplots of human vs.\ model mean ratings, and \Cref{fig:human-model-barplot} provides an aggregate bar plot comparison.

\subsection{Cross-Task Consistency}
\label{app:cross-task-consistency}

As a final probe, we examine whether model behavior is self-consistent across our three experimental formats; \Cref{tab:cross_task_consistency} reports cross-task Spearman correlations. We are particularly interested in model consistency between the multi-image generation and the dual-image rating settings: GPT-4o Mini is the least consistent ($\rho=0.29$), while Gemini-2.5 Flash is the most ($\rho=0.86$). Whether a VLM is proprietary or not has little effect on cross-task consistency, though, without access to internal model information, we cannot compare this same correlation with a reference probability approach. Overall, our three paradigms measure related, though distinct, aspects of novel reference behavior.

\begin{table}[t]
\centering
\small
\vspace{-10pt}
\setlength{\tabcolsep}{3pt}
\renewcommand{\arraystretch}{1.15}
\begin{tabular}{c|ccc}
\toprule
\textbf{Model} & \multicolumn{3}{c}{\textbf{Cross-Task Spearman} ($\rho$)} \\
 & \shortstack{\textbf{Gen.}\\$\leftrightarrow$\textbf{Likert}} & \shortstack{\textbf{Gen.}\\$\leftrightarrow$\textbf{Log Prob.}} & \shortstack{\textbf{Likert}\\$\leftrightarrow$\textbf{Log Prob.}} \\
\midrule
{\footnotesize Qwen-2 VL 7B}     & 0.45 & \textbf{0.55} & \textbf{0.85} \\
\midrule
{\footnotesize Idefics-3 8B}     & 0.36 & 0.07 & 0.23 \\
\midrule
{\footnotesize Molmo-2 8B}       & 0.35 & 0.39 & 0.40 \\
\midrule
{\footnotesize GPT-4o Mini}      & 0.29 & ---  & ---  \\
\midrule
{\footnotesize Gemini-2.5 Flash} & \textbf{0.86} & ---  & ---  \\
\bottomrule
\end{tabular}
\caption{Cross-task Spearman correlations ($\rho$) per model aggregated on all conditions; all results are significant (p\textless 0.001) except for Idefics-3 \textbf{Gen.} $\leftrightarrow$ \textbf{Log Prob}. Dashes indicate task combinations unavailable for closed-source models. Bold values mark the highest per-column.}
\label{tab:cross_task_consistency}
\vspace{-10pt}
\end{table}

\FloatBarrier
\newpage
\section{Human--Model Statistical Correlations}
\label{sec:human-model-correlations}

\Cref{tab:rho_results} reports Spearman correlations ($\rho$) between human and model Likert-scale ratings, computed at three granularities: overall (all conditions aggregated), leave-one-out (each perturbation type excluded in turn), and single perturbation type (each type in isolation).

\begin{table*}[ht]
\centering
\resizebox{\textwidth}{!}{\begin{tabular}{lcccccc}
\toprule
Subset & $n$ & Idefics3 8B & Molmo2 8B & Qwen2-VL 7B & Gemini 2.5 Flash & GPT-4o-mini \\
\midrule
\multicolumn{7}{l}{\textit{Overall}} \\
All & 56 & .742 & .771 & .812 & .905 & \textbf{.915} \\
\midrule
\multicolumn{7}{l}{\textit{Leave-One-Out (P-Type)}} \\
Part Addition & 48 & .729 & .843 & .816 & .904 & \textbf{.924} \\
Part Removal & 48 & .770 & .694 & .752 & .858 & \textbf{.899} \\
Style Degradation & 48 & .753 & .769 & .836 & .915 &  \textbf{.925} \\
Shape Deformation & 48 & .729 & .760 & .803 & .889 & \textbf{.907} \\
Color Shift & 48 & .717 & .736 & .809 & \textbf{.907} & .901 \\
Texture Shift & 48 & .688 & .763 & .824 & .904 & \textbf{.913} \\
Background & 48 & .723 & .823 & .784 & \textbf{.904} & .903 \\
\midrule
\multicolumn{7}{l}{\textit{Single P-Type}} \\
Part Addition & 8 & .012 & .690 & .802 & .667 & \textbf{.833} \\
Part Removal & 8 & .455 & .756 & .619 & .786 & \textbf{.929} \\
Style Degradation & 8 & .833 & .905 & .619 & .881 & \textbf{.929} \\
Shape Deformation & 8 & .794 & .635 & .589 & .395 & \textbf{.717} \\
Color Shift & 8 & -.521 & \textbf{.988} & -.481 & .764 & .970 \\
Texture Shift & 8 & .351 & .946 & .667 & .433 & \textbf{.958} \\ 
Background & 8 & .505 & .648 & \textbf{.750} & -.253 & .248 \\
\bottomrule
\end{tabular}}
\caption{Spearman correlation ($\rho$) across models and perturbation types.}
\label{tab:rho_results}
\end{table*}

\FloatBarrier
\newpage
\section{Additional Ablation Results and Figures}
\label{sec:additional-ablations}

We present supplementary figures for the ablation studies described in \Cref{sec:ablations}.

\subsection{Visual Similarity Ablation}
\label{app:vis-similarity-ablation}

\begin{figure*}[!t]
    \centering
    \includegraphics[width=\textwidth]{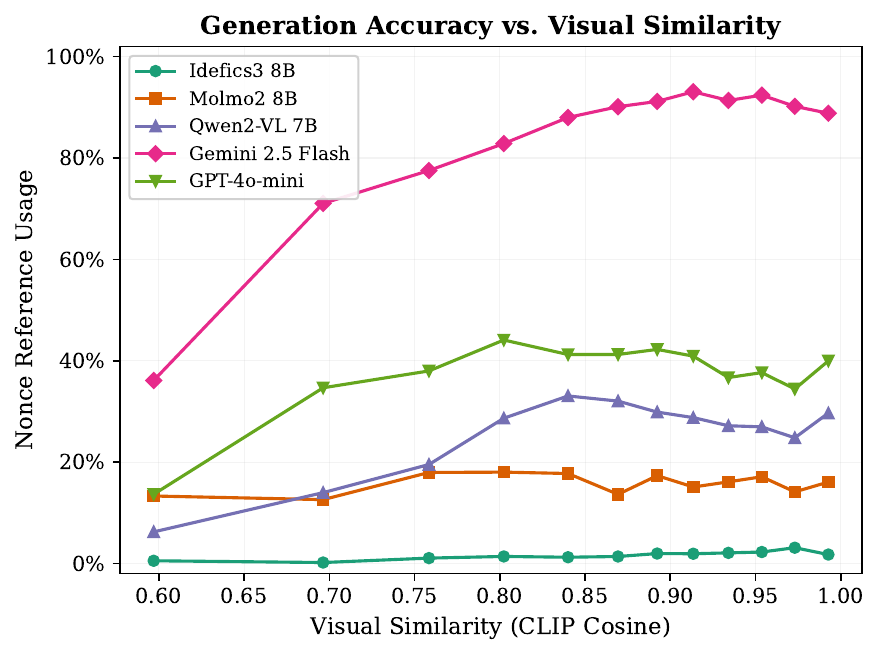}
    \caption{Model performance and ratings as a function of visual similarity between each original and perturbed image pair.}
    \label{fig:visual-sim-ablation}
\end{figure*}

\Cref{fig:visual-sim-ablation} examines how visual similarity (measured via CLIP ViT-B/32 cosine similarity) between each original and perturbed image pair relates to both generation accuracy and Likert-scale ratings. We observe a small overall increase in model ratings as similarity increases, but the effect varies sharply across models: Gemini-2.5 Flash's rating increases from 40\% on average when CLIP cosine similarity is 0.6 to 90\% when it is 1.0, while Molmo-2 only increases by 2\%.

\FloatBarrier
\subsection{In-Context Pool Composition Ablation}
\label{app:pool-composition}

\begin{figure*}[!t]
    \centering
    \includegraphics[width=\textwidth]{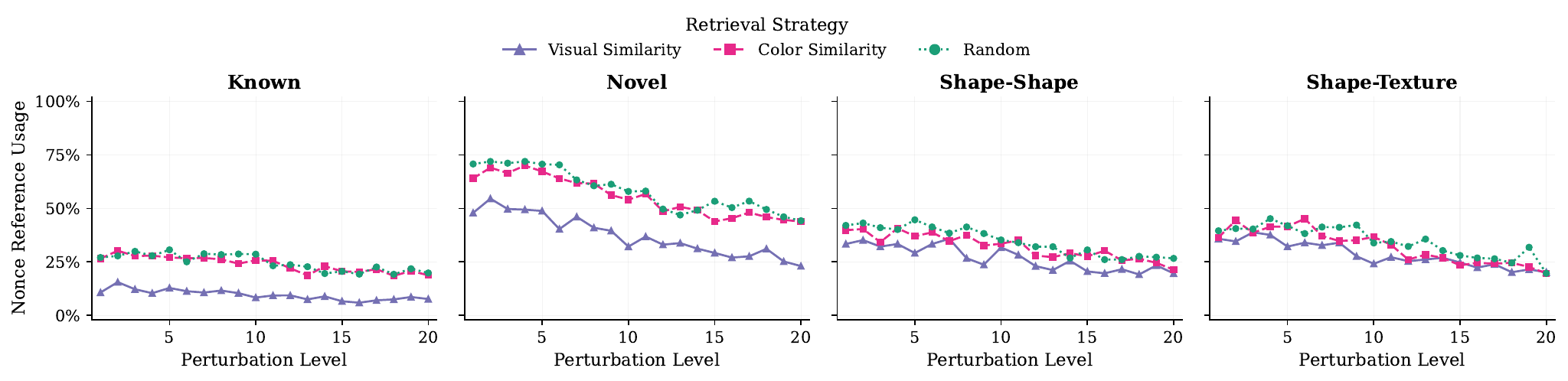}
    \caption{Qwen-2 VL 7B performance across pool composition strategies: random, color similarity, and visual similarity (CLIP).}
    \label{fig:pool-sim-ablation}
\end{figure*}

\Cref{fig:pool-sim-ablation} compares Qwen-2 VL 7B performance across three different strategies for composing the in-context image pool: visual similarity (CLIP-based, our default), color similarity (foreground-masked HSV histograms), and uniform random sampling. Random and color-similarity pools yield 6--10\% more nonce reference usage across object categories on average, confirming that the visual-similarity setup is the least trivial task for Qwen-2.

\FloatBarrier
\subsection{Prompt Agreement Ablation}
\label{app:prompt-agreement}

\begin{figure*}[!t]
    \centering
    \includegraphics[width=\textwidth]{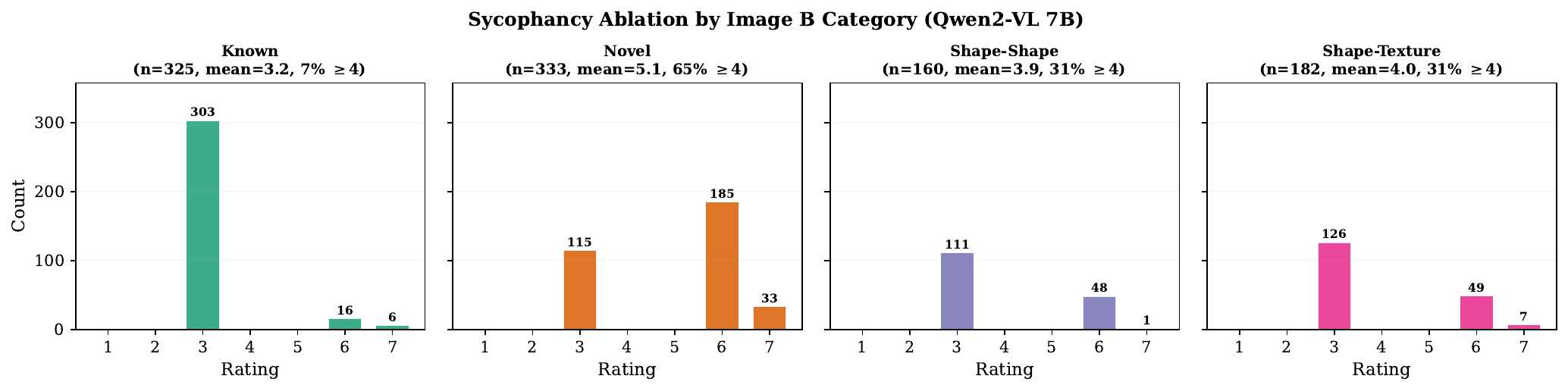}
    \caption{Qwen-2 VL 7B responses on the ablated ``failure case'' trials, by object category.}
    \label{fig:syco-obj-cat}
\end{figure*}

\begin{figure*}[!t]
    \centering
    \includegraphics[width=\textwidth]{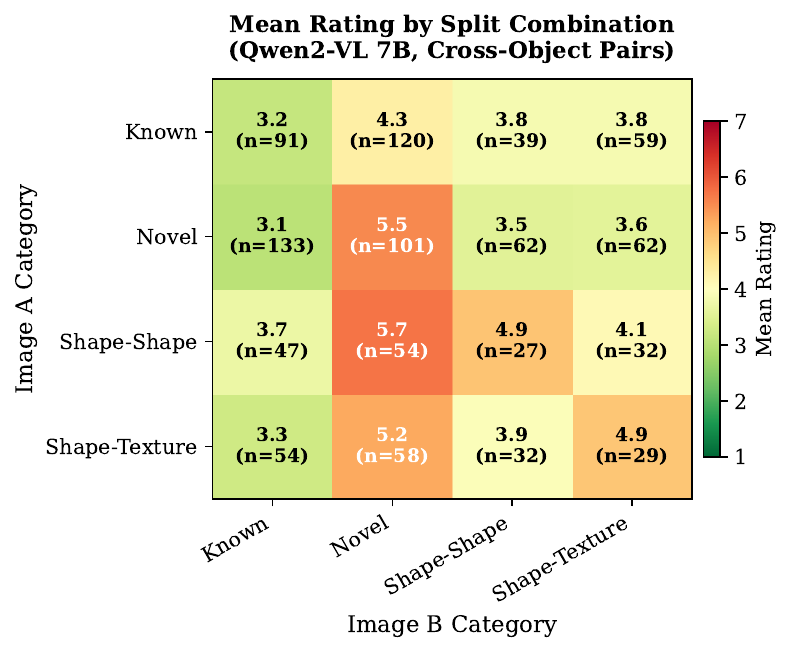}
    \caption{Heatmap of Qwen-2 VL 7B responses on the ablated ``failure case'' trials, by object category.}
    \label{fig:syco-heatmap}
\end{figure*}

\Cref{fig:syco-obj-cat,fig:syco-heatmap} show results from the prompt agreement (``sycophancy'') ablation, where image pairs in the Likert-scale rating setup are composed of images from \emph{different} objects rather than perturbations of the same concept. \Cref{fig:syco-obj-cat} shows the distribution of Qwen-2 VL 7B responses by object category, while \Cref{fig:syco-heatmap} provides a heatmap of mean ratings broken down by the object categories of both Image~A and Image~B.

\section{Behind the Scenes}
Inspired by final author BK, primary author AT would like to provide a reflection and glimpse into the work that was put in to make this project happen, with the goal of offering transparency and discussion around the project and the scientific research process as a whole.

\subsection{Formulating the Problem}
Author AT recalls a lecture in McGill University's Language Acquisition course where the Gavagai Problem was first introduced, that it should be theoretically incredibly difficult for vision-language mappings to be acquired from minimal exposures, but that humans exhibit no such difficulty due to several factors, including inductive biases like the shape bias, mutual exclusivity bias, whole object constraint, etc. AT then began to wonder: do VLMs also have these biases? Could it be that, if these biases were misaligned or absent, then this could contribute to the gap in language learning efficiency between humans and machines? While this was an optimistic hypothesis, it got the wheels turning on what would be completed and presented in this paper.

Originally, the plan was to focus in on what was happening during the model training, so we initially planned on training a VLM from scratch on some image-captioning dataset and probing responses on some curated evaluation set. In fact, unified models were even considered for the task, as they could undergo the most interesting and controllable evaluations of language acquisition from both visual and linguistic stimuli, until it was promptly understood that training a unified model from scratch on an academic research budget and resources was non-trivial. So, AI2's Molmo was chosen instead for the task, and, after a long while of setting the training up and customizing for our particular analyses, the experiments were ready to be carried out. What a relief it was to have all of the moving parts working as intended, until the results revealing themselves to be largely inconclusive across the board---there just weren't any directional patterns in how certain biases were developing during model training, not even for lower-level ones like color and texture.

As it turns out, nearly everybody the work was discussed with agreed that probing using in-context learning was the most interesting anyways, as a direct comparison with human judgments could be done. This was far more straightforward to set up and produced cleaner results that yielded intriguing conclusions about novel concept learning in VLMs on the whole.

\subsection{Final Reflections}
Similar to author BK's reflection in their most recent work, this research marks the end of a critical chapter in my (author AT's) career and life---in this case, the undergraduate journey. AT has been incredibly fortunate to have been involved in lots of academic research during the degree, an opportunity that many others don't get at this stage in their education, as well as to be supported throughout the years by an amazing supervisor, author SR, and all the extraordinary collaborators and mentors, beyond just those mentioned in this work. Research is one of the most rewarding journeys that one can experience.

\end{document}